\documentclass[journal]{IEEEtran}
\usepackage{pgfplots}
\pgfplotsset{compat=1.18}
\usepackage{amsmath,amsfonts}
\usepackage{array}
\usepackage[caption=false,font=normalsize,labelfont=sf,textfont=sf]{subfig}
\usepackage{stfloats}
\usepackage{url}
\usepackage{verbatim}
\usepackage{graphicx}
\usepackage{cite}
\usepackage{multirow}
\usepackage{float}
\usepackage{arydshln}
\usepackage{color}
\usepackage[utf8]{inputenc}
\usepackage{booktabs}
\usepackage{algorithm}
\usepackage{algpseudocode}
\usepackage{pgfplots}
\usepackage{pgfplotstable}
\usepackage{tabularx}
\usepackage{xcolor}
\usepackage[T1]{fontenc}
\usepackage{newtxtext,newtxmath}
\usepackage{overpic}
\usepackage{amsmath}
\DeclareFontShape{TS1}{cmr}{m}{sc}{
  <-> ssub * cmr/m/n
}{}

\definecolor{rblue}{rgb}{0,0.5,1}
\definecolor{hollywoodcerise}{rgb}{0.96, 0.0, 0.63}
\definecolor{lasallegreen}{rgb}{0.03, 0.47, 0.19}
\definecolor{hanpurple}{rgb}{0.32, 0.09, 0.98}
\definecolor{green(pigment)}{rgb}{0.0, 0.65, 0.31}
\usepackage[pagebackref=false,breaklinks=true,colorlinks,bookmarks=false]{hyperref}
\hypersetup{colorlinks,linkcolor={red},citecolor={hanpurple},urlcolor={magenta}} 
\newcommand{\YFF}[1]{\textcolor{black}{#1}}
\newcommand{\YF}[1]{\textcolor{black}{#1}}
\begin{document}

\title{\YF{Learning Granularity-Aware Affordances from Human-Object Interaction for Tool-Based Functional Dexterous Grasping}}
\author{Fan Yang, Wenrui Chen, Kailun Yang, Haoran Lin, Dongsheng Luo, Conghui Tang,\\Zhiyong Li, and Yaonan Wang
\thanks{This work was partially supported by the National Key R\&D Program of China under Grant 2022YFB4701400/2022YFB4701404, the National Natural Science Foundation of China under Grant 62273137, 62473139, No. U21A20518, and No. U23A20341, the Hunan Provincial Research and Development Project under Grant 2025QK3019, the Hunan Science Fund for Distinguished Young Scholars under Grant 2024JJ2027, and the Open Research Project of the State Key Laboratory of Industrial Control Technology, China (Grant No. ICT2025B20).
(\textit{Corresponding author: Wenrui Chen. E-mail: chenwenrui@hnu.edu.cn.)})}% <-this % stops a space
\thanks{F. Yang, W. Chen, K. Yang, H. Lin, D. Luo, and C. Tang are with the School of Artificial Intelligence and Robotics, Hunan University, Changsha 410012, China. (E-mail: ysyf293@hnu.edu.cn.)}
\thanks{W. Chen, K. Yang, Z. Li, and Y. Wang are also with the National Engineering Research Center of Robot Visual Perception and Control Technology, Hunan University, Changsha 410082, China.}}
\maketitle
\begin{abstract}
To enable robots to use tools, the initial step is teaching robots to employ dexterous gestures for touching specific areas precisely where tasks are performed. Affordance features of objects serve as a bridge in the functional interaction between agents and objects. However, leveraging these affordance cues to help robots achieve functional tool grasping remains unresolved. To address this, we propose a granularity-aware affordance feature extraction method for locating functional affordance areas and predicting dexterous coarse gestures. We study the intrinsic mechanisms of human tool use. On one hand, we use fine-grained affordance features of object-functional finger contact areas to locate functional affordance regions. On the other hand, we use highly activated coarse-grained affordance features in hand-object interaction regions to predict grasp gestures. Additionally, we introduce a model-based post-processing module that transforms affordance localization and gesture prediction into executable robotic actions. This forms GAAF-Dex, a complete framework that learns Granularity-Aware Affordances from human-object interaction to enable tool-based functional grasping with dexterous hands. Unlike fully-supervised methods that require extensive data annotation, we employ a weakly supervised approach to extract relevant cues from exocentric (Exo) images of hand-object interactions to supervise feature extraction in egocentric (Ego) images. To support this approach, we have constructed a small-scale dataset, \YFF{\textbf{F}unctional \textbf{A}ffordance \textbf{H}and-object Interaction Dataset (FAH)}, which includes nearly $6K$ images of functional hand-object interaction Exo images and Ego images of $18$ commonly used tools performing $6$ tasks. Extensive experiments on the dataset demonstrate that our method outperforms state-of-the-art methods, and real-world localization and grasping experiments validate the practical applicability of our approach. The source code and the established dataset are available at \url{https://github.com/yangfan293/GAAF-DEX}.
\end{abstract}


\section{Introduction}\label{int}

\IEEEPARstart{E}{nabling} robots to flexibly utilize tools based on diverse task instructions (e.g., pressing, grasping, or opening) constitutes a cornerstone of human-robot collaboration, with functional grasping~\cite{zhang2023functionalgrasp} serving as a critical initial step. Unlike general grasping, \YF{functional grasping imposes stringent requirements, necessitating dynamic identification of task-specific functional regions and generation of corresponding grasping gestures. Specifically, it entails: (1) precise localization of task-specific functional regions (e.g., drill's button or scissor's handle) rather than arbitrary contact points; (2) generation of multi-finger grasps via dexterous hands to meet the complex demands of varied tasks.} 
This work aims to achieve functional grasping through vision-guided approaches, leveraging constraints inherent to objects and tasks.

\begin{figure}[t]
\centering
\includegraphics[width=0.48\textwidth]{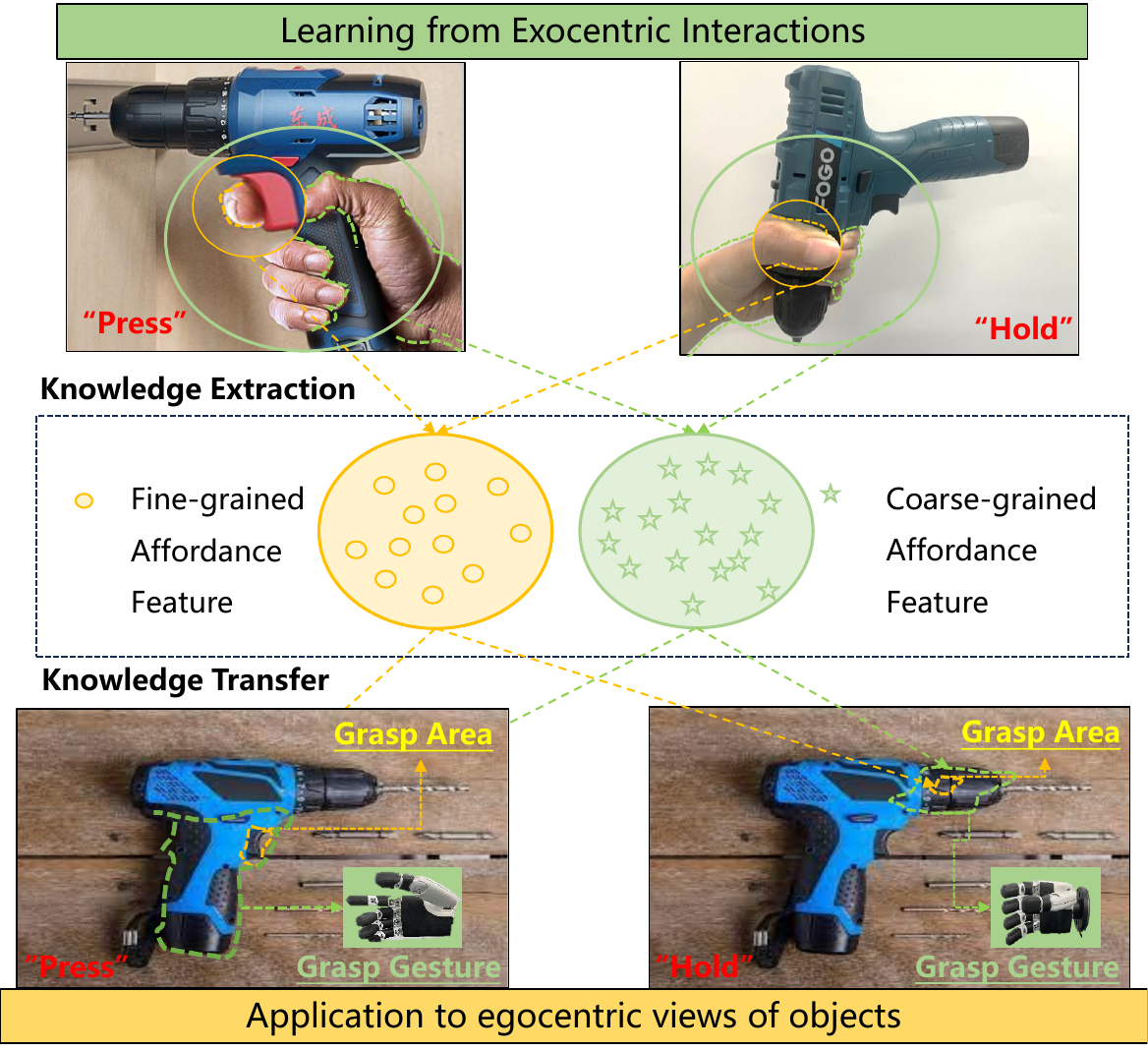}
\vskip-2ex
\caption{GAAF-Dex extracts multi-granular affordance features from exocentric images and transfers them to egocentric images, achieving functional grasping through localization, gesture prediction, and post-processing.}
\vskip-3ex
\label{fig:intro1}
\end{figure}

Conventional vision-based grasping methods primarily focus on 6D pose estimation~\cite{DBLP:journals/corr/abs-1910-11669, DBLP:journals/corr/abs-1711-00199, Yu2023CatEgoryLevel6O}, capable of determining an object's overall position and orientation, but inadequate for localizing fine-grained functional regions or predicting task-specific grasping gestures. Data-driven approaches attempt to regress hand-object interaction parameters (e.g., contact points and gestures) directly from images~\cite{corona2020ganhand, jian2023affordpose, yang2022oakink, mandikal2021learning, jiang2021hand} but rely heavily on extensive pixel-level annotations and often utilize human hand models, rendering them ill-suited for practical robotic hand applications. Deep reinforcement learning methods train grasping policies in simulation to output end-effector poses and finger configurations~\cite{wu2020generative, li2021learning, mandikal2021learning}, yet their dependence on complex simulation setups and parameter tuning limits real-world applicability.

The concept of ``affordance''~\cite{gibson1977theory} offers a novel perspective to address these shortcomings by characterizing potential physical interactions of object components (e.g., a button for pressing), thereby linking tasks to actions. Recent advances in visual affordance research~\cite{stark2008functional, kjellstrom2011visual, nagarajan2019grounded, chen2023affordance, luo2023learning} have shown promise: object-centric methods can detect, segment, and label ``action possibility'' regions on objects, while weakly supervised methods~\cite{luo2023learning, nagarajan2019grounded, li2021Ego, li2023locate, luo2022learning} learn affordance regions from exocentric (Exo) hand-object interaction images and transfer them to egocentric (Ego) perspectives, significantly reducing annotation costs. However, these approaches are limited to coarse-grained action region localization, failing to fully exploit multi-granular affordance cues in exocentric images or provide a holistic solution integrating localization and gesture prediction.

\begin{figure}[t]
\centering
\includegraphics[width=0.48\textwidth]{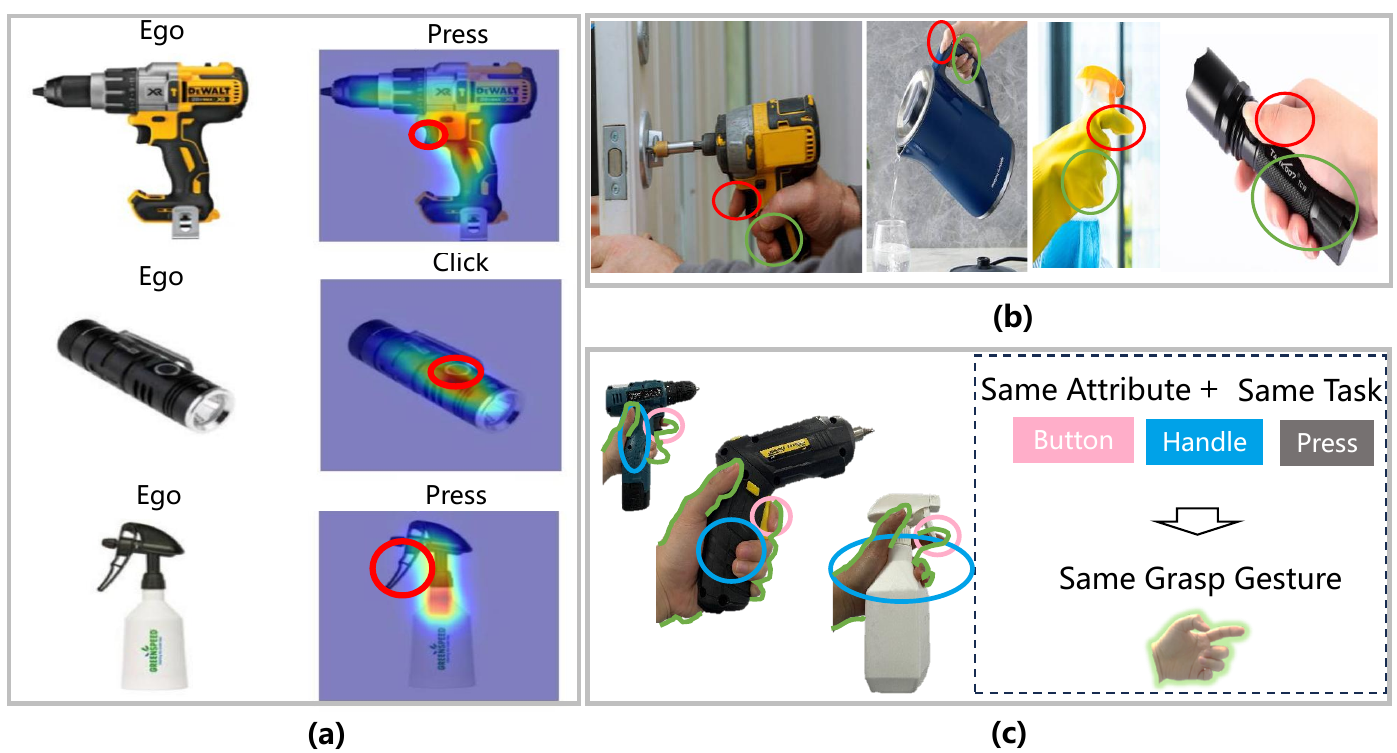}
\vskip-2ex
\caption{Research motivations. (a) Functional grasping requires fine-grained affordance localization (red circle), beyond coarse-grained action regions (colored areas). (b) Objects with similar structural attributes employ consistent gestures for identical tasks. (c) Functional fingers (red circle) exhibit spatial separation from other fingers (green circle).}
\vskip-3ex
\label{fig:intro2}
\end{figure}

\YF{To address these limitations, this paper proposes GAAF-Dex, a weakly supervised multi-task framework that leverages affordance cues from exocentric hand-object interaction images to dynamically localize functional regions and generate corresponding coarse-grained gestures based on diverse task instructions, thereby achieving functional grasping, as illustrated in Fig.~\ref{fig:intro1}, which depicts the overall pipeline from feature extraction to localization and gesture prediction. 
This approach tackles three key deficiencies of existing methods: (1) a focus solely on isolated localization tasks, neglecting multi-granular affordance cues in exocentric images; (2) detection limited to coarse regions (as shown in the colored areas in Fig.~\ref{fig:intro2} (a)), whereas functional grasping demands fine-grained localization (see the red circles in Fig.~\ref{fig:intro2} (a)); and (3) reliance of gesture prediction methods on human hand models (e.g., MANO~\cite{feix2015grasp}), which perform poorly in robotic hand applications.}

\YF{GAAF-Dex integrates a vision-driven affordance solution with multi-task learning and a post-processing module to effectively unify perception and action. Initially, a multi-task framework is designed to extract multi-granular affordance features from exocentric hand-object interaction images via weak supervision and transfer them to egocentric perspectives, providing a perceptual foundation for functional grasping. To address the precise localization demands of functional grasping (e.g., Fig.~\ref{fig:intro2} (a)'s red circle), fine-grained affordance features are extracted, coupled with spatial separation between functional fingers (Fig.~\ref{fig:intro2} (b)'s red circle) and other fingers (Fig.~\ref{fig:intro2} (b)'s green circle), employing spatial analysis and kinematic modeling to achieve functional finger-guided fine-grained localization, accurately targeting key contact points, such as a drill's button in a ``\textit{Press}'' task. Furthermore, inspired by the observation that objects with similar structural attributes (e.g., a drill and a spray bottle's button-handle design) employ consistent gestures for identical tasks (Fig.~\ref{fig:intro2} (c)), coarse-grained affordance features are extracted to predict task-specific coarse-grained grasping gestures via an affordance-driven prediction network that focuses on high-activation features in exocentric interactions, delivering diverse gestures tailored to robotic hands and overcoming limitations of human hand models. To bridge visual perception and robotic execution, a post-processing module is developed, integrating localization results and gesture predictions to compute a transformation matrix from fingertip to wrist using gesture-derived joint angles and robotic hand models, thereby executing dexterous grasping actions.}

Existing datasets, such as those proposed in ~\cite{corona2020ganhand, luo2023grounded, yang2022oakink, jian2023affordpose, zhan2024oakink2, luo2021one, hasson2019learning}, often rely on synthetic images, utilize mesh parameters for gesture representation, or lack multi-view data, rendering them inadequate for the generalization and practicality demands of functional grasping. To address this, the FAH dataset is constructed, comprising approximately $6,000$ images covering $18$ tools, $6$ functions, and $14$ gesture labels, requiring only image-level annotations to significantly reduce labeling costs while supporting task scalability and providing a practical research foundation for functional grasping.

The contributions of this work are outlined as follows:
\begin{itemize}
    \item \YF{A weakly supervised multi-task learning framework, GAAF-Dex, is proposed. It integrates fine-grained functional region localization and coarse-grained gesture prediction, leveraging multi-granular affordance features and a model-based post-processing module to bridge perception and control in a complete grasp execution pipeline.}
    \item \YF{A dataset named FAH is introduced, containing functional human-object interaction data with both region-level and gesture-level annotations. It enables affordance transfer from exocentric to egocentric views and serves as a benchmark for functional grasp learning.}
    \item \YF{The effectiveness and generalizability of GAAF-Dex are demonstrated through extensive experiments on the FAH dataset and real-world, enabling task-conditioned grasping across diverse scenarios and unseen tools.}
\end{itemize}

\section{Related work}\label{related}
\textbf{Visual Affordance Understanding.} 
Research in vision-based affordance understanding aims to locate areas of objects that are operable. Various methodologies have inferred visual affordances for simple gripper grasps~\cite{luo2023learning, chu2019learning, ardon2019learning, xu2021affordance}. 
Chen~\textit{et al.}~\cite{chen2022learning} \YFF{proposed} a framework for detecting 6-DoF task-oriented grasps, processing observed object point clouds to predict diverse grasping poses tailored for distinct tasks.
The studies~\cite{song2023learning, nguyen2023language} were conducted to generalize the robot grasping affordance areas beyond labels by incorporating large prediction models.

In contrast, works such as those in~\cite{luo2023learning, luo2023grounded, li2023locate, DBLP:journals/corr/abs-2405-05552, DBLP:conf/cvpr/DelitzasTTSPE24} explored non-robot-centric perspectives in affordance understanding. Early efforts focused on fully supervised methods that required per-pixel labeling, resulting in high data acquisition costs~\cite{luo2023grounded, luo2023leverage, chuang2018learning}.  
To address this, recent studies~\cite{luo2022learning, li2023locate, nagarajan2019grounded} proposed weakly supervised approaches, leveraging human-object interaction cues from images~\cite{li2023locate} or videos~\cite{luo2023learning} to supervise affordance learning for object-only views. For example, Locate~\cite{li2023locate} \YFF{aggregated} features from exocentric images into compact prototypes (human, object parts, and background) to supervise egocentric images, enabling identification of matching object parts. Similarly, Luo~\textit{et al.}~\cite{luo2023learning} analyzed hand positions and motions in interaction videos to obtain affordance areas for object-alone images.   
Zhang~\textit{et al.}~\cite{DBLP:journals/corr/abs-2405-05552} introduced a bidirectional progressive transformer using video data to achieve joint prediction of hand trajectories and interaction hotspots in first-person scenarios. 
\YF{The research works in~\cite{DBLP:conf/cvpr/DengXW0J21,tong2024robust, tong2024edge, 10908219} advanced the field of 3D perception. Specifically, the work in~\cite{10908219} leverages unsupervised multi-view stereo (MVS) and neural rendering to enable effective perception of 3D dense and occluded scenes, which can support robotic operations in complex environments.} 
3D AffordanceNet~\cite{DBLP:conf/cvpr/DengXW0J21} focused on recognizing 3D affordances of static objects by analyzing their shapes and features, but its reliance on synthetic datasets, lack of dynamic hand-object interactions, and the absence of functional operation tasks limit its applicability to real-world and dynamic applications.

Tool use, however, demands a combination of dexterous manipulation and functional part affordance understanding. To address this, we integrate weakly supervised affordance learning with the generation of dexterous coarse gestures. Our approach not only learns affordance localization from hand-object interactions but also predicts grasping gestures, laying the groundwork for practical tool use.

\textbf{Coarse-to-Fine Dexterous Grasping.} 
Achieving functional tool manipulation with robotic hands necessitates advanced dexterous grasping, which extends beyond basic two or three-finger grippers to involve multi-finger coordination. Previous approaches to achieving precise grasping \YFF{relied} on either model-based methods~\cite{Rodriguez2012caging, Rosales2012synthesis, El2015computing, Murray2017mathematical}, which require extensive time for object and hand modeling and suffer from poor generalization, or data-driven methods~\cite{shang2020deep, mayer2022ffhnet, wei2022dvgg, wang2023task, zhu2023toward, zhang2023functionalgrasp}, which are costly due to the need for extensive labeling of contact points and joints. 
In contrast, the coarse-to-fine approach used in~\cite{corona2020ganhand, zhang2023functionalgrasp, jian2023affordpose} \YFF{treated} the task of predicting dexterous gestures as a classification problem.
After obtaining a specific category of grasp type, fine-tuning \YFF{was performed}, simplifying the high-dimensional data prediction task. 
GanHand~\cite{corona2020ganhand} \YFF{utilized} $33$ grasp classification types of the MANO model~\cite{feix2015grasp} to generate pre-grasp postures, whereas FunctionalGrasp~\cite{zhang2023functionalgrasp} \YFF{mapped} these $33$ grasp types of MANO models to the ShadowHand robotic hand model to obtain pre-grasp postures. In contrast, we have designed a classification network for $14$ gestures of a low-cost robotic hand, leveraging the consistency of the object's ``task-affordance''. These $14$ gestures encompass the daily tool operation needs of humans.

\begin{table}[t!]
\centering
\caption{Statistics of related datasets and the proposed FAH dataset. Inter-Type: Interaction type (Ha-O: hand-object, Hu-O: human-object). Real / Syn.: Real or synthetic data. View: Perspective. Annotation: Level of annotation (Pix-Level: pixel-level, Img-Level: image-level). Hand Pose: Annotation type (Mesh: hand mesh, Angle: joint angles). Aff. Int.: Affordance interaction ($\checkmark$: yes, $\times$: no)}.
\vskip-2ex
\label{data}
\renewcommand{\arraystretch}{1.5} 
\resizebox{\linewidth}{!}{
\large
\begin{tabular}{|l|l|l|l|l|l|l|l|}
\hline
\textbf{Dataset}       & \textbf{Year} & \textbf{Inter-Type} & \textbf{Real / Syn.} & \textbf{View} & \textbf{Annotation} & \textbf{Hand Pose} & \textbf{Aff. Int} \\ \hline
ObMan~\cite{hasson2019learning}           & 2019 & Ha-O    & syn. & Exo     & Pix-Level & Mesh & $\times$ \\ \hline
YCBAfford~\cite{corona2020ganhand}       & 2020 & Ha-O    & syn. & Exo     & Pix-Level & Mesh & $\times$ \\ \hline
PAD~\cite{luo2021one}             & 2021 & Hu-O    & real & Exo-Ego & Pix-Level & $\times$ & $\checkmark$ \\ \hline
AGD20K~\cite{luo2022learning}          & 2021 & Hu-O    & real & Exo-Ego & Img-Level & $\times$ & $\checkmark$ \\ \hline
OakInk-Image~\cite{yang2022oakink}    & 2022 & Ha-O    & real & Exo     & Pix-Level & Mesh & $\checkmark$ \\ \hline
AffordPose~\cite{jian2023affordpose}      & 2023 & Ha-O    & syn. & Exo     & Pix-Level & Mesh & $\checkmark$ \\ \hline
OakInk2~\cite{zhan2024oakink2}         & 2024 & Ha-O    & real & Exo     & Pix-Level & Mesh & $\checkmark$ \\ \hline
\textbf{FAH (Ours)}     & \textbf{2024} & \textbf{Ha-O} & \textbf{real} & \textbf{Exo-Ego} & \textbf{Img-Level} & \textbf{Angle} & \textbf{$\checkmark$} \\ \hline
\end{tabular}
}
\vskip-3ex
\end{table}

\textbf{Hand-Object Interaction Datasets.}
The emergence of relevant datasets has significantly advanced the development of ``hand-object'' interactions, as shown in Tab.~\ref{data}. OakInk~\cite{yang2022oakink} \YFF{introduced} a dataset containing affordances and corresponding gesture labels for $1800$ household objects. 
\YFF{AffordPose~\cite{jian2023affordpose} introduced a synthetic dataset for fine-grained hand-object interactions based on specific object part visibility, but its reliance on labor-intensive MANO annotations~\cite{feix2015grasp} limits its applicability to real-world scenarios, as it focuses solely on static interactions without addressing functional operations or dynamic interactions critical for robotics applications.} AGD20K~\cite{luo2023learning} \YFF{focused} on inferring human intentions from support images of human-object interactions and transferring them to a set of query images. However, it \YFF{did not consider} direct hand-object interactions and \YFF{lacked} gesture annotations. 
Datasets related to visibility~\cite{corona2020ganhand, luo2023grounded, yang2022oakink, jian2023affordpose, zhan2024oakink2, luo2021one, hasson2019learning} \YFF{faced} challenges such as reliance on synthetic data, use of mesh parameters for gestures, and failure to consider human behavior in reasoning about affordance areas. 
\YF{The FAH dataset we constructed provides real paired Exo-Ego view data for learning transferable human-tool interaction knowledge, and includes coarse gesture annotations that extend affordance vision research toward robot-executable manipulation.}

\section{Problem Formulation}
\begin{figure*}[t!]
    \centering
    \includegraphics[scale=0.8]{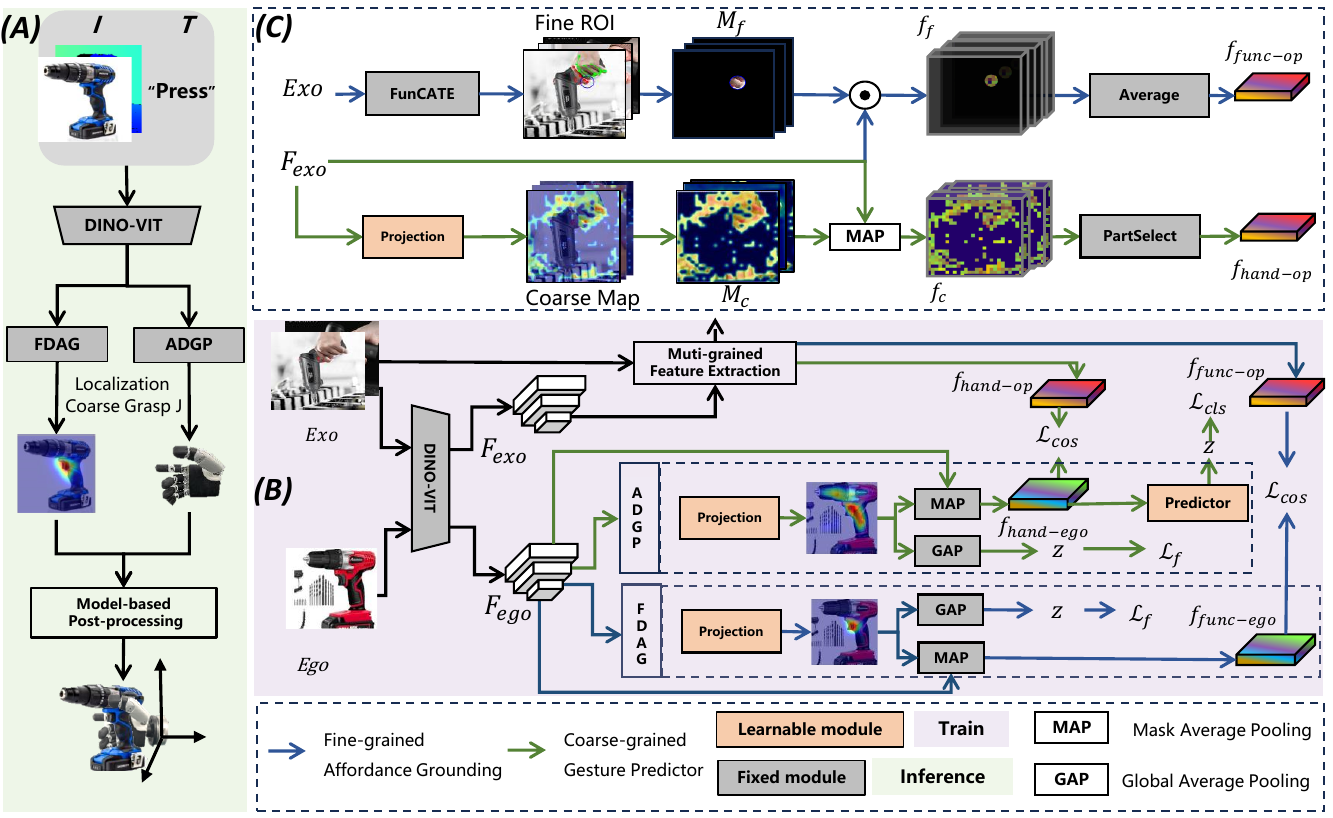}
    \vskip-2ex
\caption{\YFF{\textbf{Framework of GAAF-Dex.} (A) The inference flow. Given an RGB-D image (where the depth is not involved in training but directly provides according to the affordance location) and a task as input, the Funcfinger-Driven Affordance Grounding (FDAG) module identifies the grasping region's coordinates, and the Affordance-Driven Gesture Predictor (ADGP) module predicts the corresponding coarse gesture. Finally, the Model-based Post-processing Module integrates coordinates and gestures for final execution. (B) The training flow takes \( N \) Exo images and one Ego image as input. The bottom part of the box represents the training procedure for the ADGP and FDAG modules for Ego images. The top part describes the process of extracting coarse-grained affordance features and fine-grained affordance features from Exo images to serve as supervision for the ADGP and FDAG modules. (C) The Multi-grained Feature Extraction (MFE) module extracts coarse- and fine-grained feature prototypes from Exo for supervision.}}
\vskip-3ex
\label{PIPLINE}
\end{figure*}

The objective of this study is to address challenges in functional grasping by developing a model, denoted by $M$. This model is designed to analyze egocentric RGB images, $I$, containing a single object, along with a task description, $T$. The model outputs the initial grasping area and a coarse grasping gesture appropriate for the task. Specifically, the model predicts:
\begin{equation}
M(I, T) \rightarrow \left(P, \{\theta_1, \theta_2, \theta_3, \theta_4, \theta_5\}\right),
\end{equation}
\YF{where \(P {=} (x, y, z)\) represents the position where the object should be grasped in the camera coordinate system.} The $z$ coordinate is derived from depth maps, providing depth information about the grasping location. 
The set $\{\theta_1, \theta_2, \theta_3, \theta_4, \theta_5\}$ denotes the joint angles for a coarse grasping gesture. Upon obtaining $P$ and $\theta_i$, a post-processing module refines these predictions to determine the precise hand positions and joint angles required for effective functional grasping.

\section{Method}
Given a set of exocentric interaction images and an egocentric image of an object, our core objective is to train two prediction modules, namely the Funcfinger-Driven Affordance Grounding (FDAG) and Affordance-Driven Gesture Predictor (ADGF) modules. We extract affordance region features related to functional fingers and corresponding grasp gesture features from exocentric (Exo) images and transfer these features to egocentric (Ego) images, enabling us to locate the grasp points and gestures of functional fingers in the egocentric images. 
During the training phase, we utilize image-level affordance labels, whereas in the testing phase, the input is an egocentric image, and the outputs are the object's optimal grasp point $P$ and the associated coarse grasp gesture $G$, as shown in the green background (A) part of Fig.~\ref{PIPLINE}.

The training part of our method is illustrated in the purple background (B) part of Fig.~\ref{PIPLINE}, and the core idea is as follows: for the input images $\{I_{exo}, I_{ego}\}$ ($I_{exo} {=} \{I_1, I_2, \ldots, I_N\}$), we first use a network $\phi$ to extract deep features \YFF{$\{\mathcal{F}_{exo}, \mathcal{F}_{ego}\} {\in} \mathbb{R}^{D \times H \times W}$}. In our case, $\phi$ is a self-supervised visual transformer (DINO-ViT~\cite{caron2021emergingdinovit}), which provides excellent part-level features. Subsequently, \YFF{based on task requirements, we extract fine-grained and coarse-grained visibility cues from Exo images using the Multi-grained Feature Extraction (MFE) module (see Fig.~\ref{PIPLINE} (C)) to supervise the corresponding features extracted in Ego images by the ADGP and FDAG modules, enabling affordance localization and gesture prediction. Specifically, for functional affordance localization, as guided by the blue dashed line in Fig.~\ref{PIPLINE}, we propose a functional finger-driven fine-grained feature extraction method (Sec.~\ref{Fun-affor}). For coarse gesture prediction, as guided by the green dashed line in Fig.~\ref{PIPLINE}, we leverage the Class Activation Mapping (CAM) \cite{zhou2016learning_cam} and the hand-background-object feature prototype selection module from LOCATE~\cite{li2023locate} to extract coarse-grained features from Exo images (Sec.~\ref{handpose}).} Finally, we design a model-based post-processing module to combine functional areas and coarse grasp gestures, yielding the final end-effector grasp points and coarse-to-fine functional grasp results (Sec.~\ref{coord_cv}).

\subsection{Fine-Grained Feature Extraction for Affordance Grounding}\label{Fun-affor}
\YFF{In this section, we focus on the blue arrow flow in Fig.~\ref{PIPLINE} (B) and (C). First, the Exo images are processed through the FunCATE module in the MFE module (see Sec.~\ref{FuncExtract}) to obtain the fine-grained Region of Interest (Fine ROI). The ROI is then used to generate the fine-grained mask \( M_f \) as follows:}
\begin{equation}
\begin{aligned}
    M_f(x, y) &= 
    \begin{cases} 
    1 & \text{if } \sqrt{(x - x_0)^2 + (y - y_0)^2} \leq r, \\ 
    0 & \text{otherwise},
    \end{cases}
\end{aligned}
\end{equation}
where the mask function \YFF{\( M_f(x, y) \)} defines a circular region centered at \( (x_0, y_0) \) with radius \( r \), producing a binary mask. Simultaneously, the feature map \( \mathcal{F}_{exo} \) is upsampled as follows:
\begin{equation}
\begin{aligned}
    f_{\text{up}} &= \text{Upsample}(\mathcal{F}_{exo}),
\end{aligned}
\end{equation}
where \YFF{\( f_{\text{up}} \)} represents the upsampled feature map obtained by resizing \YFF{\( \mathcal{F}_{exo} \)} to match the size of the resized image. This ensures consistency between the feature coordinates and the ROI coordinates. \YFF{Then, \( f_{\text{up}} \) and \( M_f \) are combined via element-wise multiplication to obtain the fine-grained affordance features \( f_{{f}} {\in} \mathbb{R}^{C \times H \times W} \): \begin{equation}
\begin{aligned}
    f_{{f}} &= 
    f_{{up}} \odot M_f.
\end{aligned}
\end{equation}
Then, the \( f_{\text{fine}} \) from \( N \) images are averaged to generate the fine-grained functional affordance prototype \( f_{\text{func-op}} \) for supervision. Meanwhile, the Ego features \( \mathcal{F}_{ego} \) are processed through the FDAG module to obtain \( f_{\text{func-ego}} \), and finally, knowledge transfer from Exo to Ego is performed (see Sec.~\ref{transfer-affor})}.

\subsubsection{FunCATE}\label{FunCAT}
\begin{figure}[t!]
    \centering
\includegraphics[scale=0.18]{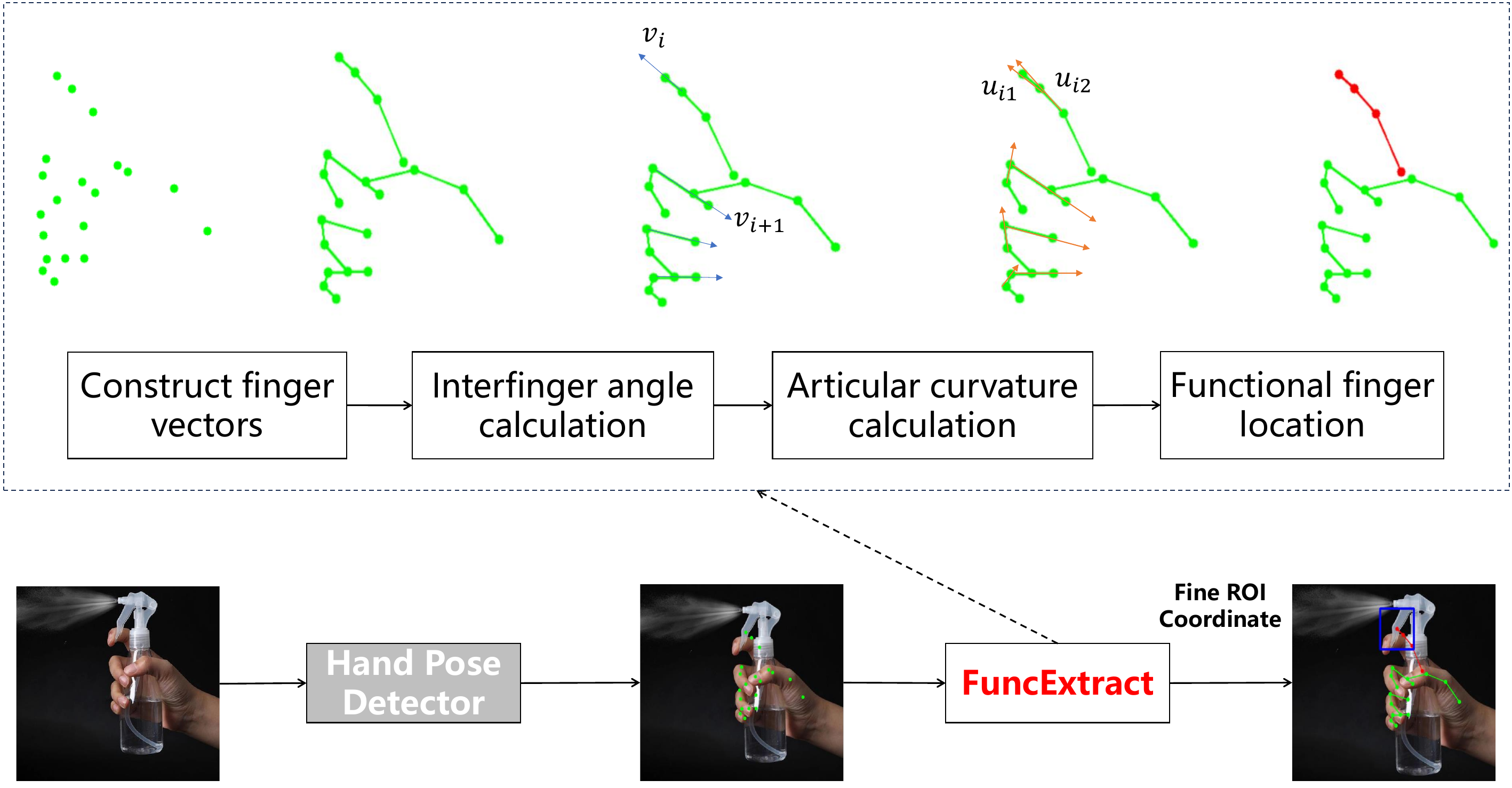}
\vskip-2ex
    \caption{FunCATE module, which includes a hand pose detector and the FunExtract module. FunExtract determines the functional finger by calculating the vector angles between fingers and the joint angles of each finger.}
    \vskip-3ex
    \label{FuncExtractp}
\end{figure}

For feature supervision, we perform functional finger-guided cue extraction on Exo images, primarily implemented by the FunCATE module. Specifically, as shown in Fig.~\ref{FuncExtractp}, we first use the network \( \phi_2 \) for gesture recognition on Exo images. In our case, \( \phi_2 \) is MediaPipe~\cite{lugaresi2019mediapipe}, which has the advantage of accurate landmark detection. 
It can obtain $21$ key points' \YFF{2D ($x$, $y$)} coordinates of the human hand. 

Then, we apply our proposed functional finger determination algorithm, FuncExtract, to obtain the \YFF{2D ($x$, $y$)}  coordinates of the functional fingertip. We consider the area with radius \( r \) around these coordinates as our fine ROI. 

\YFF{\textbf{FuncExtract:}\label{FuncExtract}  
The FunExtract module is illustrated in the dotted box at the top of Fig.~\ref{FuncExtractp}. 
This module evaluates the spatial looseness between fingers by considering both the spatial alignment (parallelism) and the local curvature (bending angle), and selects the functional finger based on this parameter.} 

Specifically, when we obtain the 2D coordinates of \(21\) key points on the human hand, we first vectorize the points of each finger and compute the cosine of the angle between adjacent finger vectors to evaluate the parallelism of the four non-thumb fingers. The cosine of the angle between vectors formed by adjacent finger joints for the \(i\)-th finger (excluding the thumb) \(\text{angle}_{i}\) is calculated as:
\begin{equation}
\text{angle}_i = \frac{v_i \cdot v_{i+1}}{\|v_i\| \|v_{i+1}\|}, \quad i \in \{2, 3, 4, 5\}.
\end{equation}
Here, \(v_i\) represents the vector of the \(i\)-th finger, \(\cdot\) denotes the dot product of the vectors, and \(\|\|\) denotes their magnitude.

\YF{If the cosine values for all adjacent finger pairs are greater than a predefined threshold \( \tau \)}, the four non-thumb fingers are considered parallel, and the thumb is directly identified as the functional finger. Otherwise, we proceed to analyze the joint bending angles of the four non-thumb fingers. 
The bending angle for each finger is determined by calculating the cosine of the angle between two vectors formed by the adjacent joints of each finger:
\begin{equation}
\begin{aligned}
\text{func}_{ID} &= \underset{i \in \{2, 3, 4, 5\}}{\mathrm{argmin}} \left( 1 - \frac{u_{i1} \cdot u_{i2}}{\|u_{i1}\| \|u_{i2}\|}\right), \\
u_{i1} &= p_{i2} - p_{i1}, \\
u_{i2} &= p_{i3} - p_{i2},
\end{aligned}
\end{equation}
where \(p_{i1}, p_{i2}\), and \(p_{i3}\) represent the coordinates of the first, second, and third joints of the \(i\)-th finger, \YF{and \(u_{i1}\), \(u_{i2}\) are vectors between the second-to-first and third-to-second joints of the \(i\)-th finger, respectively.} The finger with the minimum bending angle is selected as the functional finger, \( \text{func}_{ID}\) is the functional finger identifier, ranging from $2$ (index finger) to $5$ (little finger).

\subsubsection{Functional Part-Level Knowledge Transfer}
\label{transfer-affor}
Now we focus on the fine-grained feature extraction of Ego and use $f_\text{func-op}$ for its supervision. Specifically, in the FDAG module, we first apply the Projection function \( P() \)~\cite{li2023locate} to the ego image, which utilizes class activation mapping (CAM) techniques~\cite{zhou2016learning_cam} to generate a functionally-aware localization map, depicted as follows:
\begin{equation}
\begin{aligned}
    P_{\text{func-ego}} = P(\mathcal{F}_\text{ego} + \text{MLP}(\mathcal{F}_\text{ego})),
\end{aligned}
\end{equation}
where \( \text{MLP} \) represents a feed-forward layer, and \( P() \) consists of two \( 3{\times}3 \) convolutional layers, normalization layers, and non-linear activation functions, followed by a \( 1{\times}1 \) class-aware convolution layer. Each map \( P_c {\in} \mathbb{R}^{H \times W} \) represents the network activation for the \( c \)-th interaction.

\YFF{Then, we perform Masked Average Pooling (MAP) between the localization map and $\mathcal{F}_{\text{ego}}$, aggregating them into an embedding vector. On the other hand, a Global Average Pooling (GAP) layer is applied to the localization map to obtain the task classification scores $z$, which are used to compute the cross-entropy loss $\mathcal{L}_{{t}}$ for optimization, depicted as follows:
\begin{equation}
\begin{aligned}
f_{\text{func-ego}} = \text{MAP}(P_{\text{func-ego}}, \mathcal{F}_{\text{ego}}), \quad
z = \text{GAP}(P_{\text{func-ego}}),
\end{aligned}
\end{equation}
where the MAP operation includes a matrix multiplication between the normalized $P_{\text{func-ego}}$ and $\mathcal{F}_{\text{ego}}$.}

Finally, we use cosine loss $\mathcal{L}_{\cos}$ and concentration loss $\mathcal{L}_{\text{c}} $ to ensure the features are correctly extracted while maintaining coherence as follows:
\begin{equation}
\mathcal{L}_{\cos} = \max(1 - \frac{f_\text{func-op} \cdot f_\text{func-ego}}{\|f_\text{func-op}\|\|f_\text{func-ego}\|} - \alpha, 0),
\label{9}
\end{equation}
\begin{equation}
\mathcal{L}_{\text{c}} = \sum_{c} \sum_{u,v} \|\langle u,v \rangle - \langle \bar{u}_c, \bar{v}_c \rangle\| \cdot P_{\text{func-ego}} / z_c,
\label{10}
\end{equation}
\begin{equation}
\bar{u}_c = \sum_{u,v} u \cdot P_{\text{func-ego}} / z_c, \quad \bar{v}_c = \sum_{u,v} v \cdot P_{\text{func-ego}} / z_c,
\end{equation}
where $\alpha$ is a margin added to compensate for the domain gap as the two embeddings come from different domains. $\bar{u}_c$ and $\bar{v}_c$ represent the center of the $c$-th localization map along the $u$ and $v$ axes, and $z_c{=}\sum_{u,v} P_{\text{func-ego}}$ is a normalization term. The concentration loss forces the high activation regions of the localization maps to be close to the geometric center.

\subsection{\YFF{Coarse-Grained Feature Extraction for Gesture Predictor}}\label{handpose}
\YFF{To achieve gesture prediction, we focus on the green arrow flow in Fig.~\ref{PIPLINE} (B) and (C). For Exo images, \( \mathcal{F}_{{exo}} \) is processed through the Projection function in the Multi-grained Feature Extractor to obtain task-specific localization maps, denoted as coarse maps. These coarse maps serve as masks \( M_c \) and are processed with \( \mathcal{F}_{{exo}} \) using MAP to generate coarse-grained affordance features \( f_c \). The features from \( N \) images are concatenated and passed into the PartSelect module~\cite{li2023locate}, \YF{which clusters the object, background, and hand features within the hand-object interaction region into $K$ clusters}. Based on the Intersection Over Union (IOU) values of the similarity map with \( \mathcal{F}_{\text{ego}} \) and the saliency map obtained from Ego images processed by DINO-ViT~\cite{caron2021emergingdinovit}, the coarse-grained affordance supervision feature prototype \( P_{\text{hand-op}} \) is derived.}

For Ego images, the ADGF module first performs the same operation as the FDAG module, extracting coarse-grained hand-object interaction features $f_{\text{hand-ego}}$ through the Projection layer. $f_{\text{hand-ego}}$ is then passed through a GAP layer to obtain the task-related class label. The normalized $f_{\text{hand-ego}}$ is combined with $\mathcal{F}_{\text{ego}}$ via MAP to obtain the supervised coarse-grained affordance features.

Finally, as discussed in Sec.~\ref{int}, these coarse-grained affordance features originate from the most active regions of hand-object interactions and include affordance-guided coarse gestures. We add a coarse gesture predictor to $f_{\text{hand-ego}}$. Specifically, we use a Fully Connected (FC) classification network on $f_{\text{hand-ego}}$ to predict the grasp type $C$, classifying it into one of the $14$ grasp types that best suit the target object. This network is trained using the cross-entropy loss $\mathcal{L}_{\text{class}}$. The predicted grasp type $C$ is associated with a representative hand configuration $H_C$, consisting of the joint angles of the five fingers and the abduction angle of the thumb.

\subsection{Training Supervision} 
\label{loss}
In summary, during the training phase, the total loss consists of the following four parts:
\begin{equation}
\mathcal{L} = \mathcal{L}_{\text{cos}} + \lambda_c \mathcal{L}_c + \mathcal{L}_{\text{class}} + \mathcal{L}_t,
\end{equation}
where \(\lambda_c\) is the weight balancing these four terms, \YFF{\(\mathcal{L}_{\text{cos}}\) is defined in Eq.~\ref{9} as the cosine similarity loss between the exo coarse and fine-grained affordance feature prototypes and the ego coarse and fine-grained affordance features; \(\mathcal{L}_c\), defined in Eq.~\ref{10}, is the clustering loss; \(\mathcal{L}_t\) represents the cross-entropy loss for task classification of the Exo coarse-grained feature prototype and the Ego coarse-grained and fine-grained functional affordance feature prototypes, as described in Sec.~\ref{transfer-affor}; \(\mathcal{L}_{\text{class}}\) is the cross-entropy loss for ego gesture type prediction, as defined in Sec.~\ref{handpose}.
}

\subsection{Model-based Post-processing Module} 
\label{coord_cv}

\YFF{In this module, we first extract the top $N$ brightest RGB-D pixels from the functional affordance grounding predicted by FDAG, corresponding to the pixels with the highest probability values in the heatmap. Let the 3D coordinates of these pixels be $P_i{=}[x_i, y_i, z_i]^T$ ($i {=} 1, 2, \dots, N$), where $z_i$ is obtained from the depth camera. Then, the contact point is determined by calculating the centroid of these pixels, given by $P_{centroid} {=} \frac{1}{N} \sum_{i=1}^{N} P_i$. Finally, the contact point coordinates $P_{centroid}$ are converted to the global coordinate system using the hand-eye calibration matrix, yielding the final functional fingertip contact point $P_{wf} {=} [x_{wf}, y_{wf}, z_{wf}]^T$.}

\begin{figure}[t!]
\centerline{\includegraphics[width=0.48\textwidth]{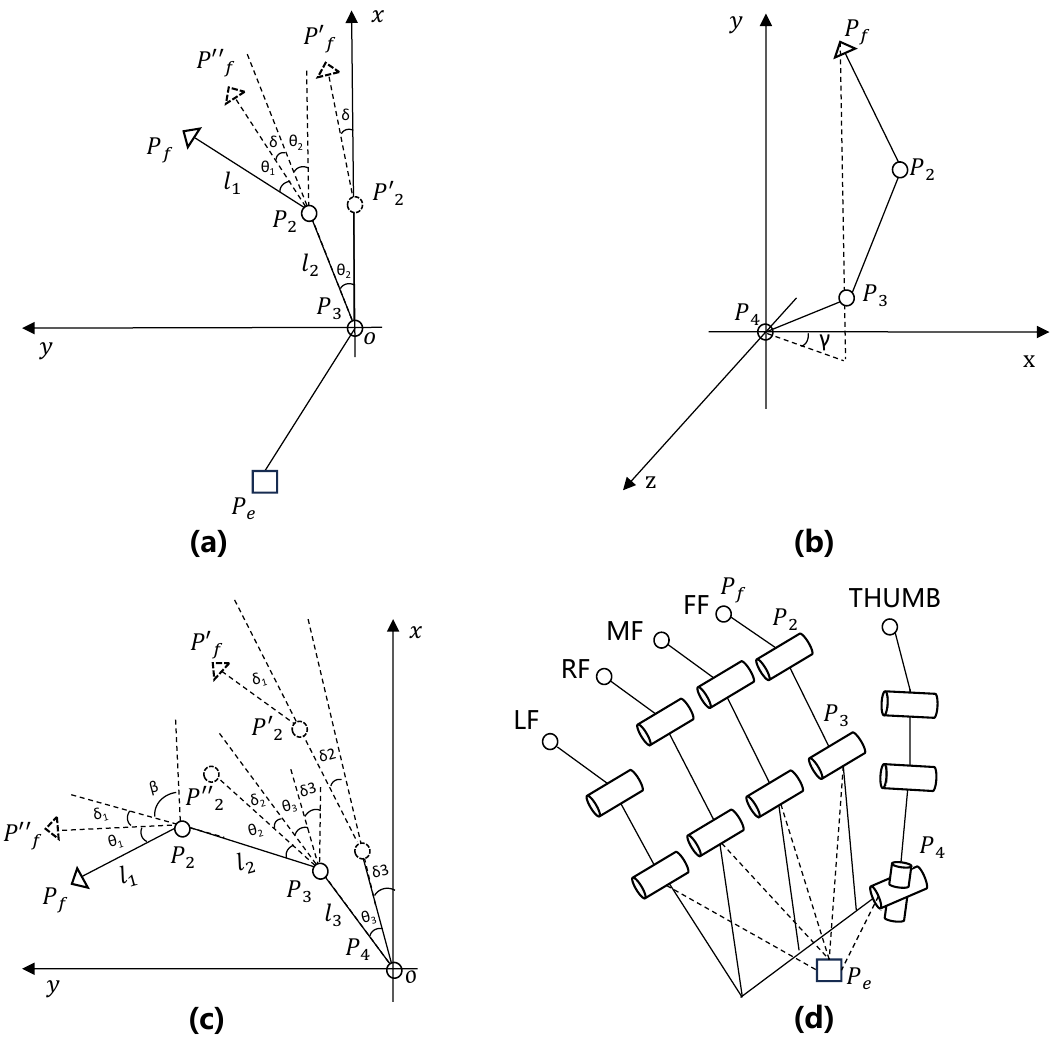}}
\vskip-2ex
\caption{Diagrams illustrating fingertip-to-end coordinate transformations based on a model. (a) shows the coordinate transformation of the flexion and extension joints of the index finger, (b) shows the coordinate transformation of the abduction and adduction joints of the thumb, (c) shows the coordinate transformation of the flexion and extension joints of the thumb, and (d) provides an overall reference for Inspire hand joints.}
\vskip-3ex
\label{model}
\end{figure}

Then, based on the proportional relationship and joint angles of the robotic hand model's finger joints, we transform the fingertip coordinates $P_{wf}$ to obtain the wrist end coordinates $P_{we}$ in the global coordinate system. Specifically, as shown in Fig.~\ref{model} (d), outside the thumb, the other four fingers of the Inspire hand have the same structure, a motor drives the two finger joints to flex and stretch. 
We take the index finger as an example as shown in Fig.~\ref{model} (a), where \( P_2, P_3 \) represent the node of the first and second finger joint rotation axes, respectively. 
$P'_i$ represents the position of the joint node when the drive motor is in the zero position. Here, \(P'_f{-}P'_2 \) is at an \(\delta \) angle to the X-axis. \( P''_f \) represents the hypothetical position of the fingertip if only the second phalanx moves. We establish a hand coordinate system with \( P_3 \) as the origin, \( O \). 
\( P_3{-}P'_2 \) is the positive direction of the x-axis, and the z-axis coincides with the rotation axis of \( P_3 \). 

The coordinate \( P_{f} \) {=} \([x_{hf}, y_{hf}, z_{hf}]^T\) of the fingertip in the hand coordinate system can be obtained as follows:
\begin{equation}
\label{fe}
P_{f}
=
R(\theta_2)
\begin{bmatrix}
l_2 \\
0 \\
0
\end{bmatrix}
+
R(\theta_1 + \theta_2 + \delta)
\begin{bmatrix}
l_1 \\
0 \\
0
\end{bmatrix},
\end{equation}
where the $l_1$ and $l_2$ represent the first and second direct lengths, respectively. $\theta_1$ and $\theta_2$ come from the linear transformation of the index finger angle of our predicted coarse gestures.
\[
R(\theta) =
\begin{bmatrix}
\cos \theta & -\sin \theta & 0 \\
\sin \theta & \cos \theta & 0 \\
0 & 0 & 1
\end{bmatrix}.
\]

The thumb is driven by a motor for flexion and extension of the three joints, as shown in Fig.~\ref{model}~(c). The calculation method of the fingertip to the end in flexion and extension is similar to that of the index finger, but the difference is that the thumb also has a lateral swing movement, as shown in Fig.~\ref{model}~(b). The mapping of the fingertip \( P_{f} {=} [x_{hf}, y_{hf}, z_{hf}]^T \) to the end in the lateral swing process is as follows:
\begin{equation}
   \ P_{f} =
\begin{bmatrix}
\cos \gamma & 0 & \sin \gamma \\
0 & 1 & 0 \\
-\sin \gamma & 0 & \cos \gamma
\end{bmatrix}
 P^o_{f},
\end{equation}
where the \(\gamma\) is the abduction-adduction angle, $P^o_{f}$ denotes the thumb fingertip coordinate in the hand coordinate system, obtained from the flexion and extension angles using a forward kinematics formulation, similar to that in Equation~\ref{fe}.

The end coordinate \( P_{we} \) in the world coordinate system can be obtained by the following equation:
\begin{equation}
P_{we} = R_{wf}(P_{e} - P_{f}) + P_{wf},
\end{equation}
where \( P_{e} {=} [x_{he}, y_{he}, z_{he}]^T \) is the wrist end coordinate in the hand coordinate system, which is directly obtained from the mechanical structure. $R_{wf}$ represents the rotation matrix of the object correctly grasped by the hand. Since this method focuses on the functional area and does not involve rotation, we assume that it is a known quantity.

Finally, to quickly and stably achieve coarse-to-fine gesture adjustment, we adopt the Functionally Integrated Adaptive Force-Feedback Manipulation (FAFM) algorithm \cite{yang2024task} to refine the coarse gesture angles. During this process, continuous force feedback is received, and the adjustment stops when the rate of change of the force derivative reaches zero, indicating a stable grasp.

\begin{table}[t]
\centering
\caption{\YFF{Definitions of six tasks in the FAH dataset with examples of target objects and corresponding contact parts for each task.}}
\vskip-2ex
\label{6tasks}
\renewcommand{\arraystretch}{1}
\setlength{\tabcolsep}{4pt} %
\begin{tabular}{>{\centering\arraybackslash}m{1cm}|>{\arraybackslash}m{5.4cm}|>{\centering\arraybackslash}m{1.5cm}}
\hline
\multirow{2}{*}{\textbf{Task}} & \multirow{2}{*}{\textbf{Definition}} & \multicolumn{1}{c}{\textbf{Example}} \\
& & \multicolumn{1}{c}{\textbf{(Object/Part)}} \\
\hline\hline
Press & \YF{Applies a gesture to a tool's button with sustained force to maintain functionality for subsequent operations.} & Drill/Button \\
\hline
Click & \YF{Applies a gesture to a tool's switch with brief force to trigger functionality for subsequent operations.} & Mouse/Switch \\
\hline
Hold & \YF{Five-finger grasp to ensure stability, facilitating subsequent actions.} & Bottle/Body \\
\hline
Open & Objects that are detached or twisted for special functionality. & Bottle/Lid \\
\hline
Clamp & \YF{Two-finger opposing force on a single region for precise control.} & Plug/Body \\
\hline
Grip & \YF{Multi-finger balance across discontinuous regions.} & Scissors/Handle \\
\hline
\end{tabular}
\vskip-3ex
\end{table}

\section{Established Dataset}

\begin{figure*}[t]
\centerline{\includegraphics[width=1\textwidth]{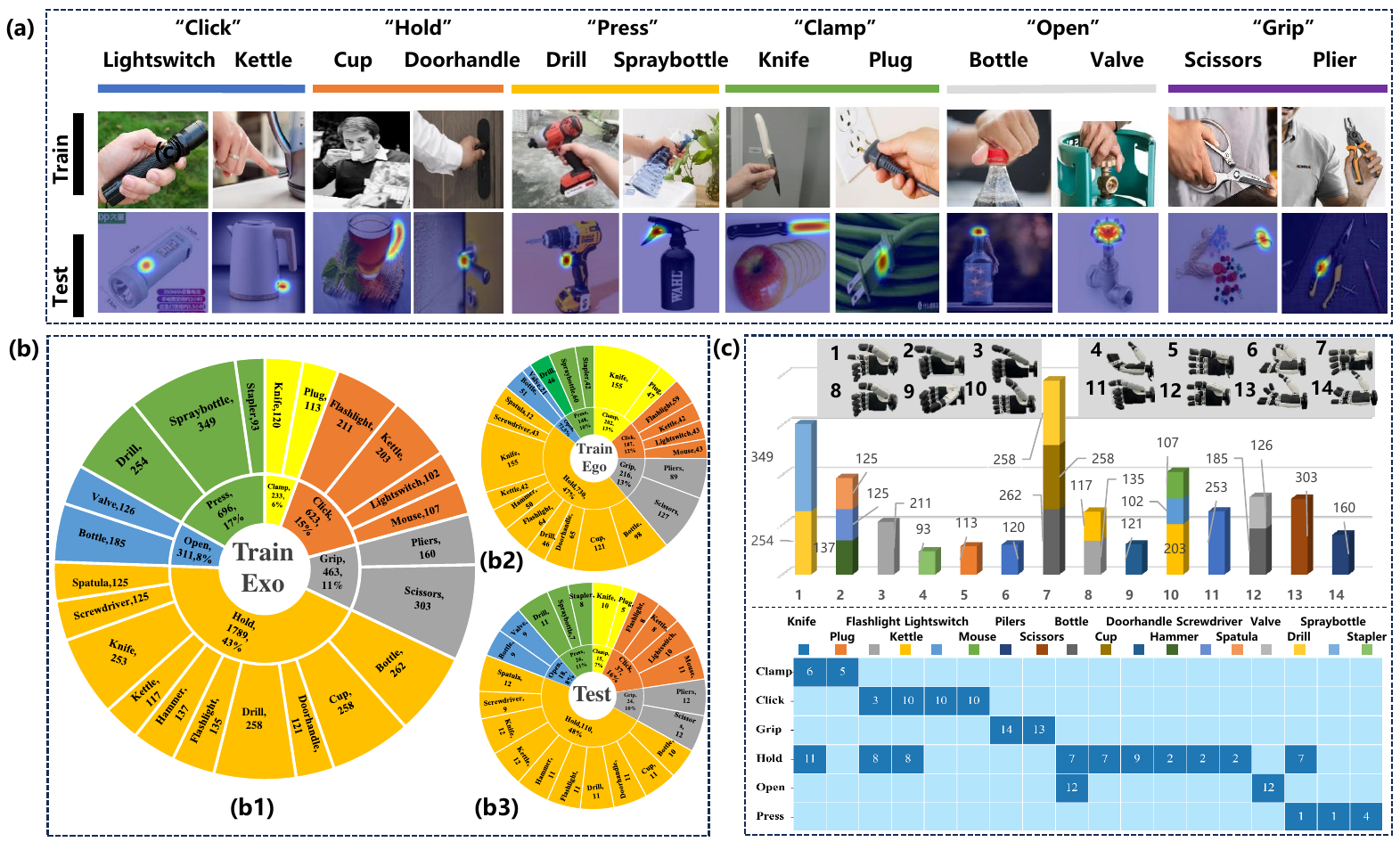}}
\vskip-2ex
\caption{\YF{Properties of the FAH dataset. (a) Examples from the dataset. (b) Instance count distribution of 18 tools and 6 tasks in the training and test sets. (c) Distribution properties of 14 coarse gestures: instance counts and their distribution across object categories (top); confusion matrix between affordance and object categories (bottom, horizontal axis: object categories, vertical axis: affordance (task) categories, table numbers: corresponding gestures).}}
\vskip-3ex
\label{dataset}
\end{figure*}

To advance research in dexterous functional manipulation, we introduce the FAH dataset, specifically designed for complex functional grasping tasks. FAH features diverse human-tool interaction examples that reflect common scenarios in both domestic and industrial settings. \YF{The dataset contains nearly $6K$ images, including 5616 training images (3951 exocentric and 1555 egocentric) and 232 egocentric test images.}

Based on the Finger-to-Function (F2F) knowledge graph~\cite{yang2024task}, FAH includes 18 commonly used tools (e.g., \textit{``Screwdriver''}, \textit{``Plug''}, \textit{``Kettle''}, \textit{``Drill''}) and 6 task types (\textit{``Press''}, \textit{``Click''}, \textit{``Hold''}, \textit{``Open''}, \textit{``Clamp''}, \textit{``Grip''}), as defined in Table~\ref{6tasks}.

\subsection{Image Collection}

Exocentric images were collected from three main sources: the AGD20K dataset~\cite{luo2022learning}, high-resolution product images from e-commerce platforms, and publicly available images retrieved using object-related keywords. To supplement underrepresented interaction types, we recruited 10 volunteers to photograph themselves using tools in natural hand-tool interactions, covering cases like \textit{``Clamp Knife''}, \textit{``Click Kettle''}, \textit{``Click Mouse''}, \textit{``Hold Drill''}, and \textit{``Open Valve''}.

A key design decision was to focus on single-hand exocentric images, which are essential for learning precise functional interaction cues. Multi-hand images were excluded to avoid interference in fine-grained affordance feature extraction. For egocentric images-where no human-object interaction is present—we directly selected standalone tool images from the same high-quality sources. Examples are shown in Fig.~\ref{dataset} (a).

Fig.~\ref{dataset} (b1), (b2), and (b3) illustrate the instance distributions across the exocentric train set, egocentric train set, and egocentric test set. The distributions are highly consistent, with the \textit{``Hold''} task being the most common in all sets (43\% in the exocentric set), while \textit{``Clamp''} and \textit{``Grip''} are less frequent (6\% and 11\% respectively). Each task-tool pair in the exocentric train set contains at least 100 images. The most common pair, \textit{``Press Spraybottle''}, includes 349 images. These statistics reflect a balanced and comprehensive dataset design that captures both common and nuanced interactions.

\subsection{Data Annotation}

\textbf{Image-level Annotation:}  
\YF{We annotated each training image with task and object labels. Exocentric images were labeled based on observed human-object interactions, while egocentric images-without interactions-were assigned task labels by mapping from tool categories, and object labels were directly annotated. Two annotators labeled independently, with a third resolving disagreements.} The task-tool relationship is many-to-many, totaling 23 combinations. For example, the task ``\textit{Hold}'' applies to tools like ``\textit{Knife}'', ``\textit{Flashlight}'', ``\textit{Cup}'', and ``\textit{Door Handle}'', while a single tool may map to multiple affordances, such as ``\textit{Knife}'' with ``\textit{Hold}'' and ``\textit{Clamp}'', and ``\textit{Flashlight}'' with ``\textit{Hold}'' and ``\textit{Click}''.

\textbf{Affordance Annotation:}  
For test images, we adopted a heatmap-based annotation approach similar to AGD20K. Annotators focused on marking the contact areas of ``functional fingers''. Three volunteers used the LabelMe tool to draw polygons over expected interaction regions based on typical ``\textit{Task Tool}'' usage. The annotations were averaged and smoothed using a Gaussian blur.

\textbf{Coarse Gesture Annotation:}  
\YF{Following F2F~\cite{yang2024task}, we assigned one of $14$ coarse grasping gestures to each ``\textit{Task Tool}'' pair in the FAH dataset. Their distributions across object categories are shown at the top of Fig.~\ref{dataset} (c), whereas the bottom shows a confusion matrix mapping gestures to task and object categories.} For each gesture, we recorded five-finger flexion angles and thumb abduction angles. Importantly, our method is robot-agnostic and can be adapted to different robotic hands by adjusting gesture parameters accordingly.

\section{Experiments}
We evaluate our approach on three levels: (1) qualitative and quantitative assessment of affordance localization based on functional fingers; (2) validation of affordance-based coarse gesture prediction; and (3) real-world dexterous grasping experiments with fixed rotation to verify localization, gesture prediction, and overall grasp success.

\begin{figure*}[t!]
\centering
\includegraphics[width=\textwidth]{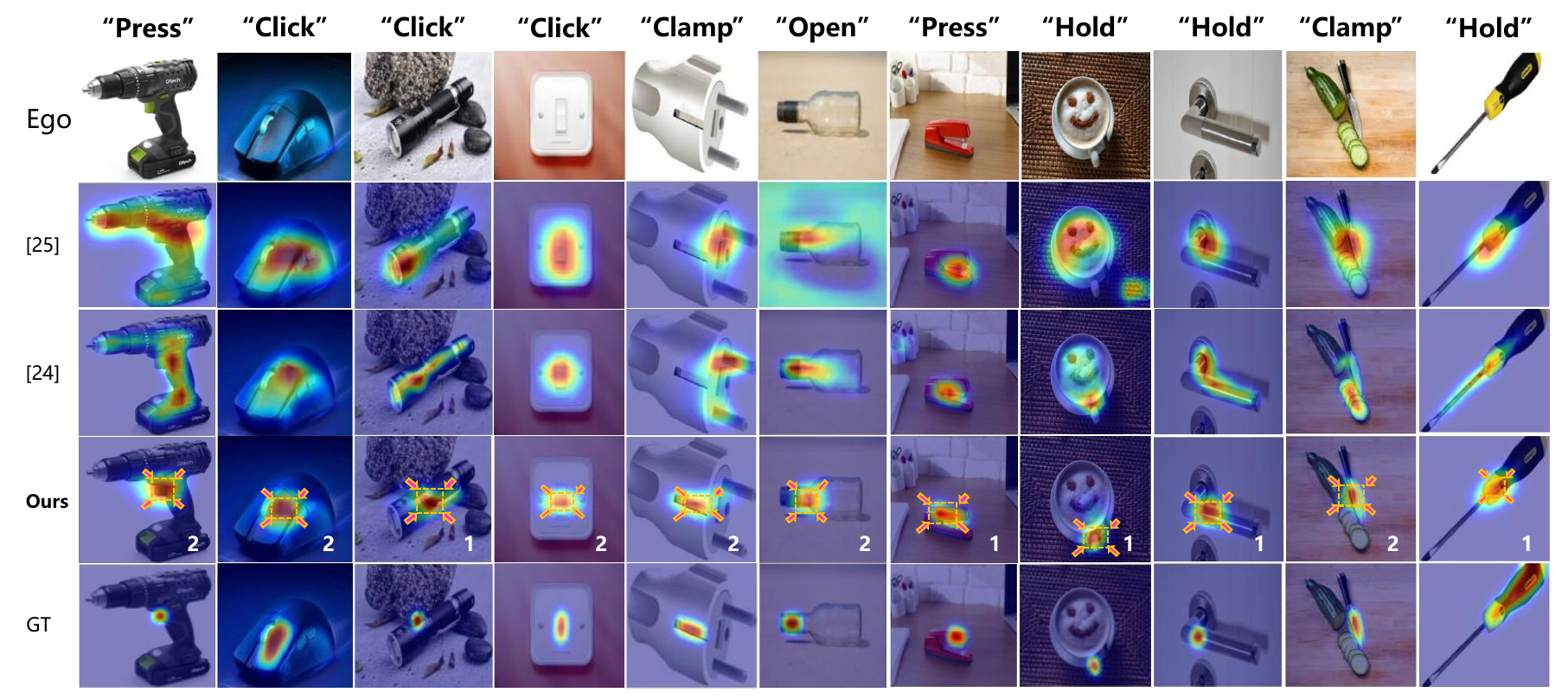}
\vskip-2ex
\caption{Qualitative comparison of our method with LOCATE~\cite{li2023locate} and Cross-view-AG~\cite{luo2022learning} on the FAH test set. The digits ``1'' and ``2'' in the fourth row of each image represent the functional finger indices, as calculated by Sec.~\ref{FuncExtract}, where ``1'' denotes the thumb and ``2'' denotes the index finger.}
\vskip-1ex
\label{vs1}
\end{figure*}

\subsection{Setups}
\textbf{Implementation Details.} 
We use the DINO-ViT-S~\cite{caron2021emergingdinovit} pretrained on ImageNet~\cite{deng2009imagenet} (unsupervised) with a patch size of $16$ to extract deep features. Each training iteration inputs one egocentric and $N{=}3$ exocentric images. Images are resized to $512{\times}512$, randomly cropped to $448{\times}448$, and horizontally flipped. We train with SGD (lr=$1e{-}3$, weight decay=$5e{-}4$, batch size=$16$). The loss weight $\lambda_c$ is set to $0.07$, and the margin $\alpha$ to $0.5$. In the first epoch, $L_{\text{cos}}$ is disabled to avoid supervision from inaccurate initial localizations.

\textbf{Metrics.} For affordance grounding, we adopt Kullback-Leibler Divergence (KLD), Similarity (SIM), and Normalized Scanpath Saliency (NSS) following prior work~\cite{li2023locate, luo2022learning}.

\YF{For gesture prediction, the accuracy for tool \( j \) in task \( t \), \( A_{j,t} = \frac{C_{j,t}}{N_{j,t}} \), where \( C_{j,t} \) is the number of correct predictions and \( N_{j,t} \) is the total number of samples. The overall average accuracy is \( AA = \frac{\sum_{t,j} C_{j,t}}{\sum_{t,j} N_{j,t}} \).}

For dexterous grasping, we measure the grasp success rate as defined in~\cite{zhu2023toward}: a grasp is successful if the hand holds the object stably for at least ten seconds and correctly performs the intended action on the tool's functional area.

\subsection{Results of Functional Affordance Grounding}
In this section, we present both qualitative and quantitative results to demonstrate the effectiveness and efficiency of our proposed method \YFF{on the FAH test set}. Our baselines include \YFF{two weakly supervised methods, Cross-view-AG~\cite{luo2022learning} and LOCATE~\cite{li2023locate}, and two fully supervised methods, PSPNet~\cite{DBLP:conf/cvpr/ZhaoSQWJ17} and DeepLabv3~\cite{DBLP:journals/corr/ChenPSA17}}.

\begin{figure}[t]
    \centering
\includegraphics[width=0.48\textwidth]{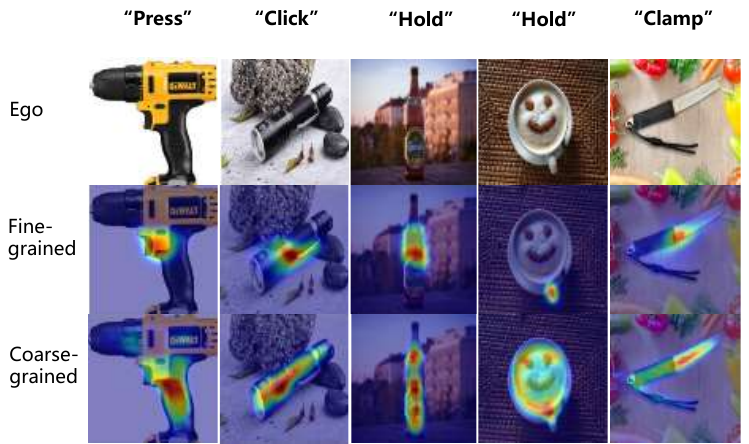}
\vskip-2ex
    \caption{Visualization of fine-grained and coarse-grained affordance feature regions. The second row shows fine-grained affordance regions used to localize functional contact areas between fingers and tools, while the third row depicts coarse-grained affordance regions for predicting rough grasp gestures.}
    \vskip-3ex
    \label{coarse-fine}
\end{figure}

\textbf{Qualitative Analysis.} We present the visibility grounding visualizations of two weakly supervised baseline methods, our method, and the Ground Truth (GT). As shown in Fig.~\ref{vs1}, our visibility localization is more concentrated in the areas where functional fingers should contact, compared to the two baseline methods. We highlighted our method's predictions with pink dashed boxes, which are essential for dexterous manipulation oriented toward functional usage. When the robot performs the corresponding actions in the \textit{``Task Tool''} scenarios, stricter functional area localization is required. Particularly for tools like drills and flashlights that have specific buttons, corresponding to the first and third images in the fourth row of Fig.~\ref{vs1}, our method accurately localizes to smaller button areas. For tools without buttons, our method also successfully localizes to the areas consistent with the human functional fingers. For instance, in the \textit{``Clamp Plug''} task, the localization is on the right side of the plug's head, which is the area the index finger needs to contact. 

\begin{figure*}[t!]
\centering
\includegraphics[width=\textwidth]{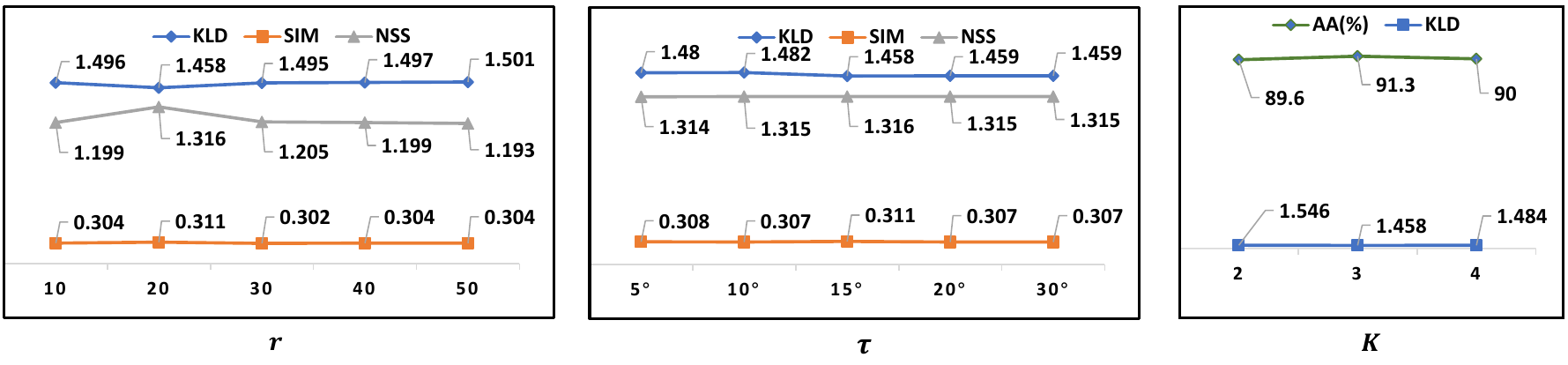}
\vskip-2ex
\caption{\YF{Hyper-parameter study. We investigate the influence of $r$ in the FunCATE module, the threshold \( \tau \) in the FuncExtract module, and the number of clusters $K$ in the PartSelect module, respectively.}}
\label{hyper}
\end{figure*}

In Fig.~\ref{coarse-fine}, we compare fine-grained and coarse-grained affordance feature extraction regions. The second row highlights the fine-grained regions utilized for precise hand-object contact localization, while the third row illustrates the larger coarse-grained regions designed for gesture prediction. The comparison demonstrates that our DAAF-Dex network effectively extracts affordance features tailored for distinct functionalities, namely contact region localization and coarse gesture prediction. For instance, in the \textit{``Press Drill''} task (first column), the button region (pressed by the index finger) in the second row represents the fine-grained feature extraction region, whereas the handle region in the third row serves as the coarse-grained feature extraction region for a full-hand grasp.

\begin{table}[t]
\centering
\vskip-2ex
\caption{Comparison to state-of-the-art weakly supervised methods and fully supervised methods (\textasteriskcentered) on the FAH test set. The \textbf{best} and \underline{second-best} results are highlighted in bold and underlined, respectively. The inference time is evaluated on a 4060Ti GPU. (↑/↓ means higher/lower is better).}
\label{pre_table}
\begin{tabular}{@{}lcccc@{}}
\toprule
Model                             & KLD ($\downarrow$) & SIM ($\uparrow$) & NSS ($\uparrow$) & Time (s) ($\downarrow$) \\ \midrule
PSPNet~\cite{DBLP:conf/cvpr/ZhaoSQWJ17}\textasteriskcentered       & 6.876              & 0.186            & 0.467            & 0.2160                  \\
DeepLabv3~\cite{DBLP:journals/corr/ChenPSA17}\textasteriskcentered    & 2.226              & 0.184            & 0.252            & 0.0540                  \\
Cross-view-AG~\cite{luo2022learning} & 1.695              & 0.269            & 1.124            & \underline{0.0226}     \\
LOCATE~\cite{li2023locate}        & \underline{1.537}  & \textbf{0.317}    & \underline{1.131} & \textbf{0.0221}        \\
Ours                              & \textbf{1.458}     & \underline{0.311}& \textbf{1.316}   & 0.0228                 \\ \bottomrule
\end{tabular}
\vskip-3ex
\end{table}

\textbf{Quantitative results.} We present the performance of the latest methods from related tasks, which involve weakly supervised object localization. As shown in Tab.~\ref{pre_table}, our method demonstrates significant improvements over competing methods across most metrics. Specifically, our approach achieves a $5.1\%$ improvement in KLD and a $16.3\%$ improvement in NSS over the state-of-the-art grounding method LOCATE~\cite{li2023locate}. Our SIM score of $0.311$ is slightly lower than LOCATE's score of $0.06$. This minor reduction is because SIM focuses more on the similarity of overlapping regions rather than their size. While LOCATE also performs part-level detection, it heavily relies on pre-trained DINO-ViT~\cite{caron2021emergingdinovit} for extracting part-level features, which are then clustered into background, human, and object categories, often neglecting the decoupling of fine-grained, object-related part features. In contrast, our approach, by localizing functional fingers during hand-object interactions, extracts more detailed functional part-level features of objects, enhancing the precision of our localization and facilitating a deeper and more effective transfer of knowledge, leading to superior performance in other metrics.

Additionally, we compare the inference time on a single image, as shown in the Time column of Tab.~\ref{pre_table}. All three methods demonstrate excellent inference speed, with times around $0.02s$. Although our method is $0.0002s$ slower than the fastest method, LOCATE~\cite{li2023locate}, this slight difference is negligible in practical applications. Moreover, unlike LOCATE~\cite{li2023locate} and Cross-view-AG~\cite{luo2022learning}, which only perform the localization task, our method also achieves coarse gesture prediction. 

To further validate the performance of our method, we compared it with two fully supervised segmentation methods, PSPNet~\cite{DBLP:conf/cvpr/ZhaoSQWJ17} and DeepLabv3~\cite{DBLP:journals/corr/ChenPSA17}. Since these methods rely on pixel-level annotations, we trained them using $1,665$ pixel-annotated Ego images from the FAH training set. The results demonstrate that our method significantly outperforms these fully supervised methods in both KLD and NSS metrics. Moreover, the inference time of our method is only $0.0228s$, which is much faster than that of PSPNet~\cite{DBLP:conf/cvpr/ZhaoSQWJ17} ($0.216s$) and DeepLabv3~\cite{DBLP:journals/corr/ChenPSA17} ($0.054s$). These results indicate that fully supervised methods perform poorly on small and imbalanced datasets, while our weakly supervised approach achieves superior performance with reduced annotation costs and faster inference efficiency.

\begin{table*}[t!]
\centering
\vskip-2ex
\caption{\YF{Accuracy of predicted grasping gestures for $6$ tasks and $18$ tools. (flashlight: fl, hammer: hm, kettle: kt, spatula: sp, scissors: sc, cup: cp, doorhandle: dh, bottle: bt, knife: kn, screwdriver: sd, drill: dr, stapler: st, spraybottle: sb, lightswitch: ls, mouse: ms, plug: pg, pliers: pl, valve: vl, average precision: ap)}}
\label{FRASP}
\begin{tabularx}{\textwidth}{lXXXXXXXXXXXXXXXXXXXX}
\toprule
 & FL. & HM. & KT. & SP. & SC. & CP. & DH. & BT. & KN. & SD. & DR. & ST. & SB. & LS. & MS. & PG. & PL. & VL. & AA\\
\midrule
Hold & 81.82 & 90.91 & 58.33 & 50 & - & 100 & 90.91 & 100 & 100 & 100 & 100 & - & - & - & - & -& - & - & 86.37 \\
Press & - & - & - & - & - & - & - & - & - & - &100& 87.5 & 100 & -& - & - & - & - & 96.15\\
Click & 75 & - & 100& - & - & - & - & - & - & - & - & - & - & 100 & 100 & - & - & - & 94.59 \\
Clamp & - & - & - & - & - & - & - & - & 100 & - & - & - & - & - & - & 80 & -& - & 93.33 \\
Grip & - & - & - & - & 100 & - & - & - & - & - & - & - & - & - & - & - & 91.67 & - & 95.83\\
Open & - & - & - & - & - & - & - & 100 & - & - & - & - & - & - & - & - & - & 100 & 100 \\
AA & 78.95 & 90.91 & 75 & 50 & 100 & 100 & 90.91 & 100 & 100 & 100 & 100 & 87.5 & 100 & 100 & 100 & 80 & 91.67 & 100 & 91.3 \\
\bottomrule
\end{tabularx}
\vskip-3ex
\end{table*}

\begin{figure*}[t!]
\centerline{\includegraphics[width=0.8\textwidth]{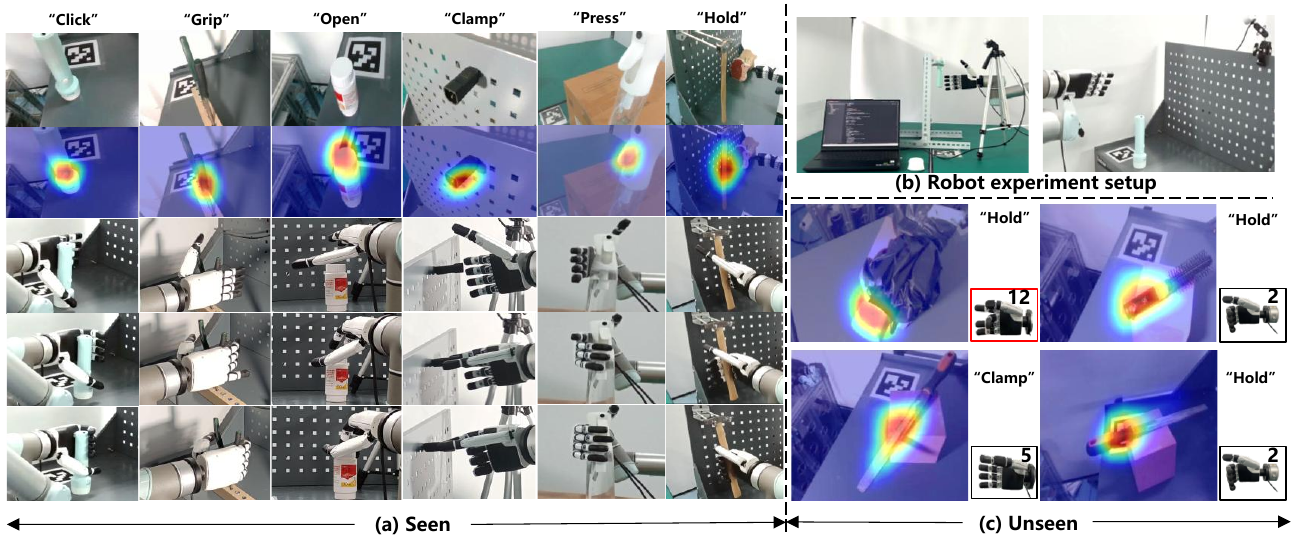}}
\vskip-2ex
\caption{\YF{Experiments in real-world scenarios: (a) Seen categories (rows 1-5: Ego image from camera view, affordance localization, approach based on localization, coarse grasping, fine grasping); (b) Hardware setup with a tool rack (left) and natural placement (right); (c) Four representative unseen categories.}}
\label{real2}
\end{figure*}

\textbf{Hyperparameter Analysis.} \YF{We further investigate the impact of the parameter \( r \) in the FunCATE module (Fig.~\ref{hyper}, left), the threshold \( \tau \) in the FuncExtract module (Fig.~\ref{hyper}, middle), and the number of clusters \( K \) in the PartSelect module (Fig.~\ref{hyper}, right). It can be observed that the threshold \( \tau \) has no significant impact on the results. Parameters \( r \) and \( K \) are respectively used to extract fine-grained and coarse-grained interaction features from exocentric data, aiding in affordance localization and gesture prediction. Their final results align with the principles of our algorithm design: an overly large \( r \) captures excessive background noise, while an overly small \( r \) fails to fully capture tool button features due to finger occlusion. When \( K = 3 \), gesture prediction accuracy reaches its highest, as more precise clustering based on the three semantic features-human, object, and background-effectively captures object features.}

\subsection{Result of Coarse Gesture Predictor}

\YF{Table~\ref{FRASP} presents the grasping gesture prediction accuracy for six tasks and 18 tools, analyzed from the following three perspectives:}

\YF{\textbf{Overall Accuracy:} The average accuracy across all task-tool combinations is $91.3\%$, indicating high prediction reliability. Tasks such as \textit{``Hold Cup''}, \textit{``Hold Bottle''}, and \textit{``Open Bottle''} achieve $100\%$ accuracy, while \textit{``Hold Spatula''} is the lowest at $50\%$, likely due to indistinct handle features and limited training data ($125$ samples).}

\YF{\textbf{Average Accuracy per Task:} All six tasks achieve over $85\%$ average accuracy. Notably, the \textit{``Hold''} task, which spans 11 tools, reaches $86.37\%$, demonstrating the model's ability to extract shared features for dexterous manipulation.}

\YF{\textbf{Average Accuracy per Tool:} Tool-wise accuracy varies. \textit{``Cup''} and \textit{``Bottle''} achieve $100\%$, while \textit{``Spatula''} and \textit{``Kettle''} yield $50\%$ and $75\%$, respectively. These lower scores are attributed to subtle and less distinctive interaction features, which challenge the gesture prediction module.}

\subsection{\YFF{Performance on Common Everyday Tools}}
\YF{We conducted experimental validation on the FAH dataset in real-world scenarios, including both unseen scenes of seen categories and unseen categories. The hardware setup for the experimental scenarios is shown in Fig.~\ref{real2} (b): the left side depicts a scene with tools suspended on a tool rack, while the right side shows a more natural placement scenario. Both setups include an Inspire Hand, a UR5 industrial robotic arm, an Intel RealSense D435i camera, a tool holder, and a control computer. The Inspire Hand, a cost-effective anthropomorphic manipulator, features six degrees of freedom: two for the thumb and one for each of the other fingers. Each degree of freedom is driven by a linear motor.}

\begin{figure}[t!]
\centerline{\includegraphics[width=0.48\textwidth]{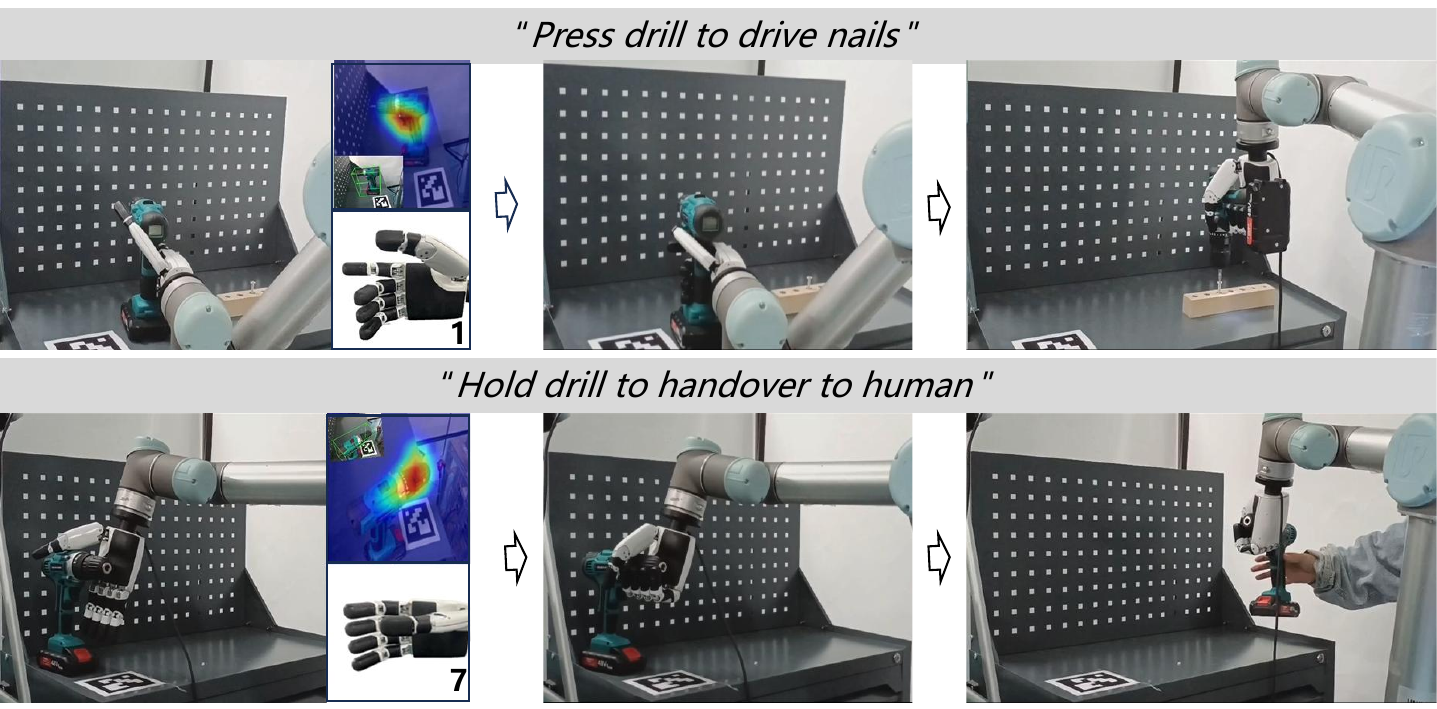}}
\vskip-2ex
\caption{\YF{Our method can dynamically adapt to subsequent different functional manipulations (the green stereoscopic frame is obtained using the 6D pose algorithm \cite{wen2024foundationpose}).}}
\vskip-3ex
\label{dyna}
\end{figure}

\YF{We first demonstrate the complete process-from localization to pre-grasping to functional grasping-on seen categories (unseen instances) in the FAH dataset. As shown in Fig.~\ref{real2} (a), for different tool instances across our six defined tasks, our algorithm accurately localizes functional regions and predicts corresponding coarse gestures, achieving functional grasping via a post-processing module.}

\YF{Our method also exhibits generalization on unseen categories, as shown in Fig.~\ref{real2} (c). For the affordance prediction task, all four unseen task-tool combinations successfully localize functional regions, such as the handle for ``\textit{Hold Umbrella}'' and the grip for ``\textit{Hold Comb}.'' For the gesture prediction task, except for ``\textit{Hold Umbrella},'' the model accurately predicts reasonable gestures for different task-tool combinations. However, ``\textit{Hold Umbrella}'' is incorrectly predicted as a ``\textit{Clamp}''-related gesture (highlighted by the red box in Fig.~\ref{real2} (c)), indicating that our gesture prediction network design requires further improvement. For the same tool ``\textit{Rasp},'' our algorithm successfully predicts distinct localizations and gestures based on varying tasks, as shown in the second row of Fig.~\ref{real2} (c).}

\YF{Furthermore, we showcase our method's dynamic adaptability for subsequent functional operations. As shown in Fig.~\ref{dyna}, for the same tool, different affordance instructions guide distinct localizations and gestures: ``\textit{Press Drill}'' can facilitate subsequent ``nailing,'' while ``\textit{Hold Drill}'' can support ``handing to a person.''}

\begin{figure}[h]
\centerline{\includegraphics[width=0.48\textwidth]{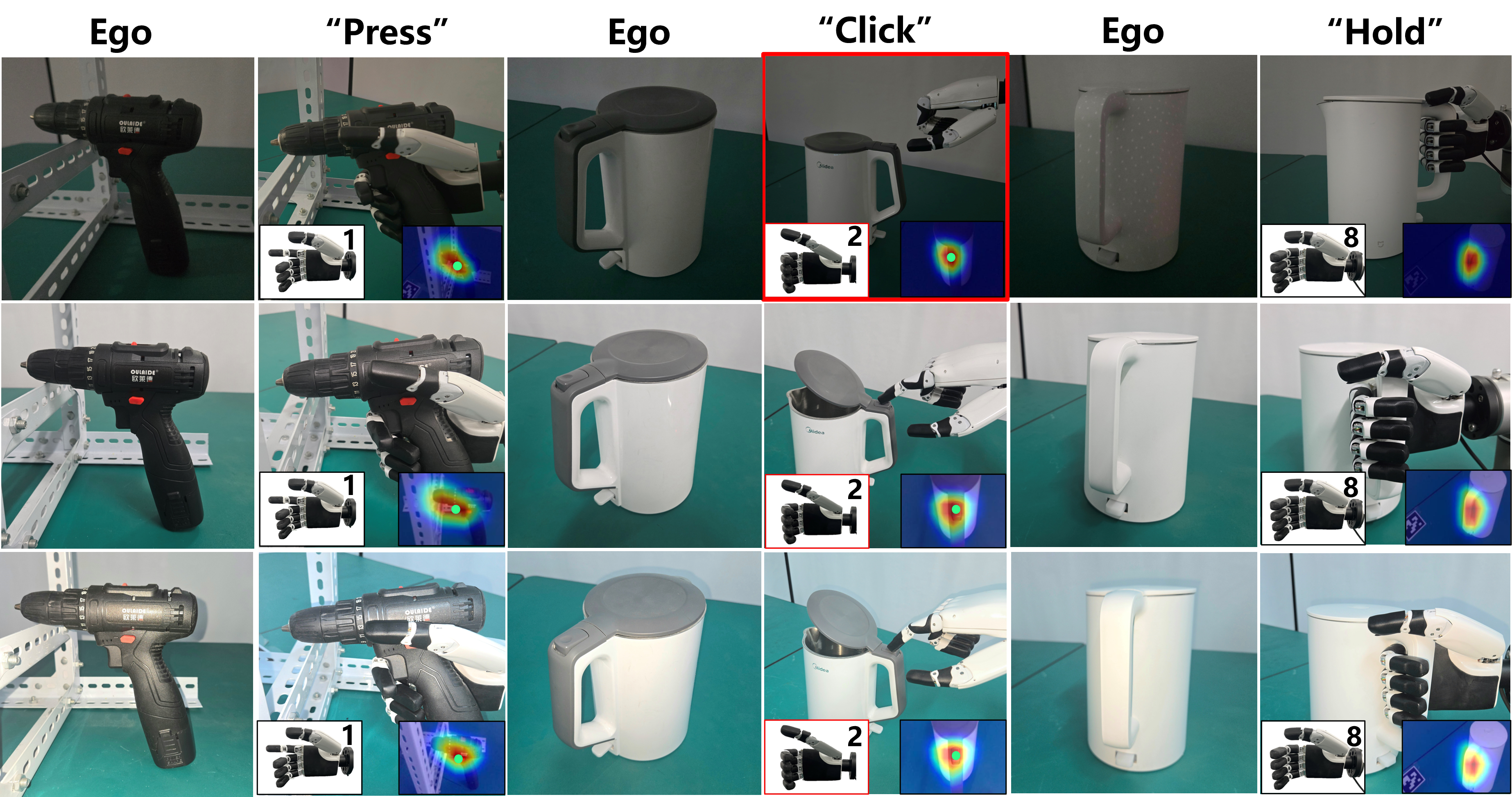}}
\caption{\YFF{Experimental results under different lighting conditions for various tools and tasks: Row 1 - dim; Row 2 - normal; Row 3 - bright. In images from even-numbered columns, the bottom left corner shows the predicted coarse hand gesture category, while the bottom right corner indicates affordance localization. Green dots represent functional finger contact points. Red boxes highlight prediction or grasping failure cases.}}
\label{light}
\end{figure}

\YFF{Secondly, our method demonstrated robustness to lighting variations. As shown in Fig.~\ref{light}, whether for the complex button-pressing task ``\textit{Press Drill}'' (columns 1 and 2) or the simpler ``\textit{Hold Kettle}'' task (columns 5 and 6), our approach consistently predicted correct coarse grasping gestures and affordance localizations under dim, normal, and bright lighting conditions. For the task ``\textit{Click Kettle},'' the functional affordance contact region—the kettle's switch—was correctly localized in all three lighting conditions (see the bottom-right corner of column 4). However, the coarse gesture category was incorrectly predicted as ``\textit{type2}'' instead of the correct ``\textit{type10}'' in all lighting conditions. Remarkably, despite the incorrect coarse gesture prediction, the kettle lid was successfully opened under normal and bright lighting conditions. This success is attributed to our functional finger determination module and the model-based post-processing module, which allowed the functional finger (index finger) to accurately interact with the switch even under the incorrect ``\textit{type2}'' gesture. This demonstrates the framework's tolerance for process errors during functional operations.}

\begin{table}[t!]
\centering
\vskip-2ex
\caption{The success rate of the representative \textit{``Task Tool''} across $15$ trials in real-world experiments. (Positioning Success Rate: Pos., Coarse Gesture Prediction Success Rate: CG., Functional Grasp Success Rate: FG., \YFF{Task Completion Time: TCT in seconds}.)}
\label{rt}
\begin{tabularx}{\linewidth}{X|X|X|X|X|X|X}
\hline
 & Press DR. & Hold DR. & Hold KT. & Click FL. & Hold HM. & Press SB. \\ \hline
Pos.& 66.67 & 13.33 & 86.67 & 66.67 & 93.33 & 93.33 \\ \hline
CG. & 46.67 & 93.33 & 100 & 66.67 & 93.33 & 46.67 \\ \hline
FG. & 26.67 & 40 & 86.67 & 66.67 & 73.33 & 46.67 \\ \hline
\YFF{TCT} & \YFF{14} & \YFF{13} & \YFF{13} & \YFF{13} & \YFF{12} & \YFF{12} \\ \hline
\end{tabularx}
\vskip-3ex
\end{table}

We also recorded the success rates of localization, coarse gesture prediction, and functional grasping, \YFF{as well as task completion times}, for $6$ representative \textit{``Task-Tool''} combinations across $15$ real-world experiments. As shown in Tab.~\ref{rt}, except for \textit{``Hold Drill''}, all other localization success rates exceeded $50\%$. The localization success rate for \textit{``Hold Drill''} was only $13\%$, attributed to limitations of the backbone model DINO-ViT~\cite{caron2021emergingdinovit}. This model provides part-level features but struggles to effectively extract features from the drill head, which lacks part-level characteristics. 

Regarding coarse gesture prediction success rates, \textit{``Hold''} tasks exhibited high success rates exceeding $93.33\%$, while other tasks showed relatively lower rates. For functional grasping success rates, we observed that, despite occasional errors in localization or coarse gesture prediction, functional grasping could still be completed. For instance, although the localization success rate for \textit{``Hold Drill''} was only $13.33\%$, the grasping task could still be successfully completed as precise localization is less critical for grasping the drill. Conversely, for \textit{``Task Tool''} combinations with high localization and coarse gesture prediction success rates, occasional lower functional grasping success rates were observed. For example, the \textit{``Press Drill''} task requires precise pressing of the button with the functional finger, posing significant challenges for selecting the end-effector grasping point. Although our model-based coordinate transformation method achieved some success, error propagation prevented precise localization.

Lastly, we recorded the average Task Completion Times (TCT) for six ``\textit{Task Tool}'' combinations, from model inference to functional grasping completion, as shown in the last column of Tab.~\ref{rt}. The task completion times ranged from $12s$ to $14s$, demonstrating relatively stable efficiency across different tasks. The small variation of $1{\sim}2s$ was primarily caused by force feedback-driven adjustments during the transition from coarse to fine-grained grasping. These consistently stable task completion times across diverse task-tool combinations highlight the robustness of our method in adapting to various task scenarios.

\begin{figure}[t!]
\centerline{\includegraphics[width=0.48\textwidth]{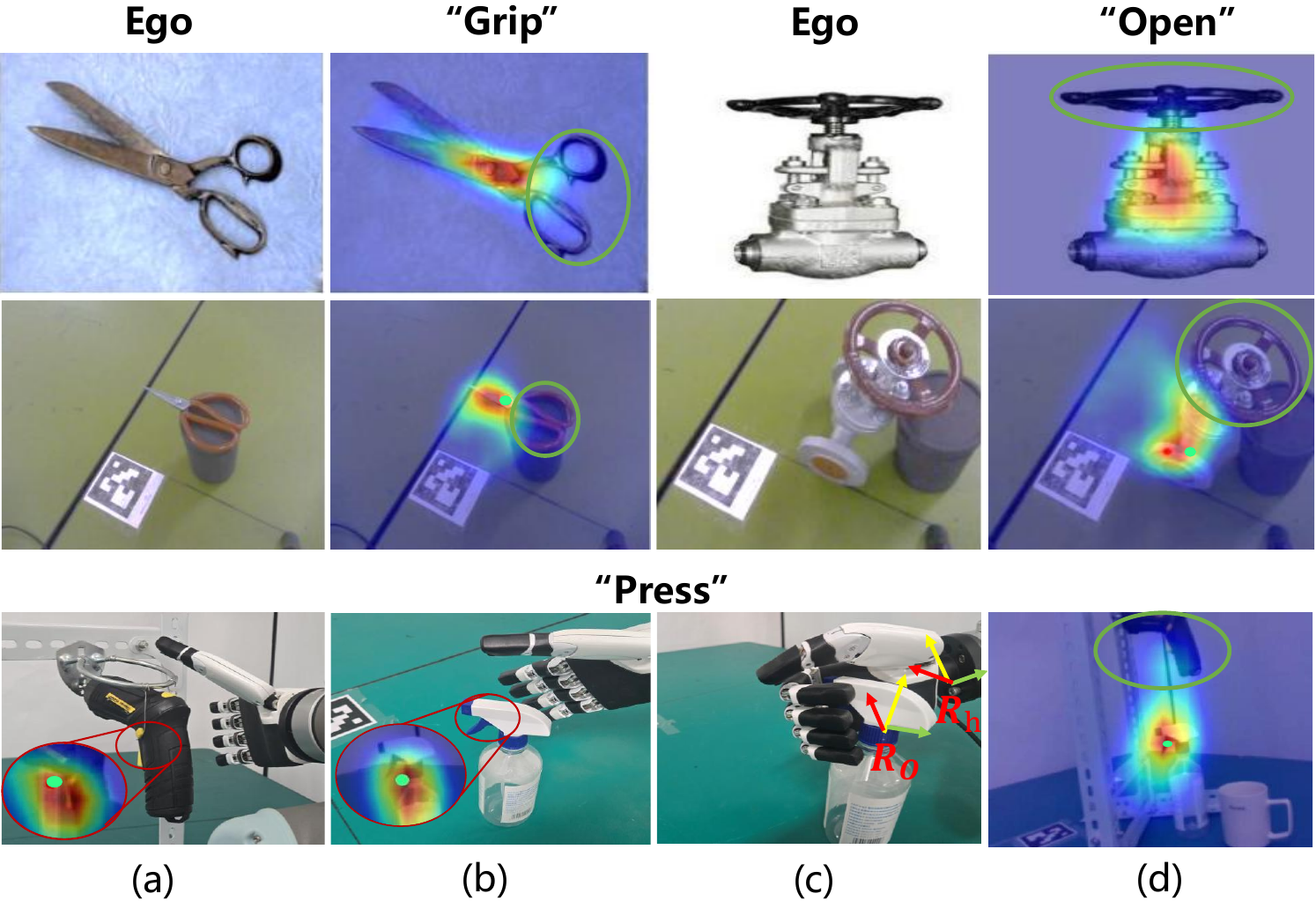}}
\vskip-2ex
\caption{\YFF{Presentation of failed cases. The green circle represents the area that should be located, and green dots represent functional finger contact points based on the affordance grounding.}}
\vskip-3ex
\label{fail}
\end{figure}

\section{Conclusion and Discussion}
In this work, we propose a weakly supervised method to learn affordance cues from exocentric images of hand-object interactions, which are used to supervise corresponding features in Ego images containing only objects. This enables the localization of functional grasping areas and coarse grasp gestures. Additionally, a model-based post-processing module refines these localizations and gestures to determine wrist-end grasp points and adjust grasps from coarse to fine, ensuring functional grasping conditions are met.

Despite the effectiveness of our method in perception-to-control functional grasping, challenges remain. Fig.~\ref{fail} highlights failure cases from the data set (first row) and real-world scenarios (second row), showing similar errors. For \textit{``Open Valve''}, localization was below the valve. These errors likely arise from functional finger features in Exo training images that overlap with incorrect regions, suggesting the need to optimize feature selection for complex tool manipulation tasks. In the third row of Fig.~\ref{fail}, failure cases in the ``\textit{Press}'' task across different tools and scenarios are presented. In (a) and (b), the affordance grounding in the RGB images was generally accurate, but depth extraction failed due to background inclusion. For example, in (a), the extracted depth corresponds to the tool rack, as indicated by the green dots. 
To address this issue, we plan to improve the localization capability in 3D environments. (c) illustrates a failure caused by the inconsistency between the initial rotation of the hand \(R_h\) and the rotation of the object \(R_o\). 
We aim to solve this problem by incorporating rotational affordance. (d) highlights the challenge of object recognition in complex scenes. In a multi-object scenario, we intended to grasp the ``\textit{Drill}'' within the green box but mistakenly localized on the ``\textit{Spraybottle}.'' 
To address this issue, we plan to leverage the multimodal alignment capability of Vision-Language Models (VLMs) to align features from natural language task instructions with those of target objects in the image, enhancing object identification and localization in complex scenes.

In summary, as one of the earliest works to integrate affordance perception with practical dexterous grasping, our method holds significant real-world value. We present a task-oriented perception-action framework with important applications in various domains. It can enable assistive robots to handle surgical tools in healthcare, support industrial robots in assembly tasks, and facilitate domestic robots in unstructured environments. Our modular, hardware-agnostic approach is adaptable to various robotic platforms and can be enhanced with multimodal data, making it applicable across industries such as agriculture, logistics, and space exploration.

\bibliographystyle{IEEEtran}   

\bibliography{resources/refs}    
\end{document}